\documentclass{article} 
\usepackage{iclr2026_conference,times}

\usepackage{amsmath,amsfonts,bm}

\def\eqref#1{equation~\ref{#1}}

\def\1{\bm{1}}

\DeclareMathAlphabet{\mathsfit}{\encodingdefault}{\sfdefault}{m}{sl}
\SetMathAlphabet{\mathsfit}{bold}{\encodingdefault}{\sfdefault}{bx}{n}

\usepackage{hyperref,xspace}
\usepackage{url}

\usepackage{graphicx}
\usepackage{wrapfig}
\usepackage[dvipsnames,table,xcdraw]{xcolor}

\usepackage[utf8]{inputenc} 
\usepackage[T1]{fontenc}    
\usepackage{hyperref}       
\usepackage{url}            
\usepackage{booktabs}       
\usepackage{amsmath,amsfonts,amssymb}       
\usepackage{nicefrac}       
\usepackage{microtype}      
\usepackage{xcolor}         
\usepackage{bm}
\usepackage{wrapfig}
\usepackage{xspace}
\usepackage{enumitem}

\newcommand{\modelname}{\textsc{QLIP}\xspace} 
\newcommand{\paperpar}[1]{\vspace{-0.75em}\paragraph*{#1}}

\newcommand{\cls}{\texttt{[CLS]}\xspace}

\newcommand{\green}[1]{{\color{Green}{#1}}}
\newcommand{\red}[1]{{\color{Red}{#1}}}

\definecolor{codegreen}{rgb}{0,0.6,0}
\definecolor{codegray}{rgb}{0.7,0.7,0.7}
\definecolor{codepurple}{rgb}{0.58,0,0.82}
\definecolor{backcolour}{rgb}{0.95,0.95,0.92}
\definecolor{darkpink}{rgb}{0.8, 0.1, 0.5}
\definecolor{almond}{rgb}{0.94, 0.87, 0.8}
\definecolor{grannysmithapple}{rgb}{0.66, 0.89, 0.63}
\definecolor{mossgreen}{rgb}{0.68, 0.87, 0.68}
\definecolor{pearl}{rgb}{0.94, 0.92, 0.84}
\definecolor{eggshell}{rgb}{0.94, 0.92, 0.84}

\title{\modelname: A Dynamic Quadtree Vision Prior Enhances MLLM Performance Without Retraining}

\author{
Kyle R. Chickering, Bangzheng Li, \& Muhao Chen \\
\texttt{\{krchicke,bzhli,muhchen\}@ucdavis.edu}
}

\iclrfinalcopy 
\begin{document}

\maketitle

\begin{abstract}
Multimodal Large Language Models (MLLMs) encode images into visual tokens, aligning visual and textual signals within a shared latent space to facilitate cross-modal representation learning. The CLIP model 
    is a widely adopted foundational vision language model whose vision encoder has played a critical role in the development of MLLMs such as LLaVA. 
    However, the CLIP vision encoder suffers from notable limitations including being constrained to only handling fixed input resolutions and a failure to produce separated embeddings for dissimilar images. 
    Replacing the vision encoder of an existing model typically incurs substantial computational costs because such a change often necessitates retraining the entire model pipeline.
    
    In this work, we identify two factors which underlie the limitations of the CLIP vision encoder: \textbf{mesoscopic bias} and \textbf{interpolation bias}. To address these issues, we propose \modelname, a lightweight adaptation of CLIP that can be seamlessly integrated with existing MLLMs with only a few lines of code and can enhance both coarse-grained and fine-grained visual understanding, without retraining the vision encoder or LLM weights. \modelname is designed around an image quadtree which replaces the standard uniform grid patches with a novel content aware patchification.
    Our experimental results demonstrate that \modelname improves the general visual question answering accuracy of the LLaVA-1.5 model series across various model sizes—without requiring retraining of the vision encoder or LLM. Notably, \modelname boosts detailed understanding performance on the challenging $V^*$ benchmark by up to 13.6\%. Code is available at \url{https://github.com/KyroChi/qlip}.
\end{abstract}

\begin{figure}[h]
    \centering
    \includegraphics[width=0.8\linewidth]{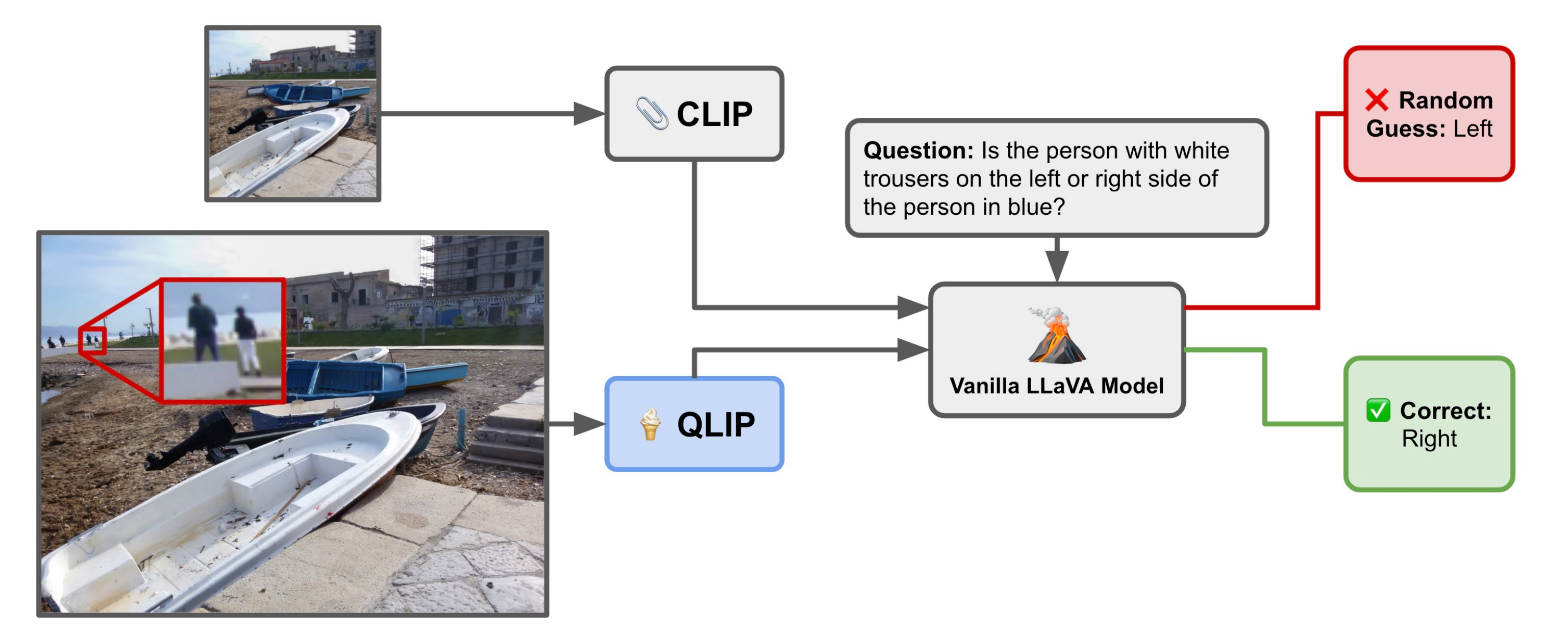}
    \vspace{-0.5em}
    \caption{\modelname is a \textbf{lightweight adaptation} of CLIP which allows models like LLaVA to perform inference on arbitrarily large images. In our experiments we find that vanilla LLaVA + \modelname gives \green{\textbf{+13.6\%}} accuracy on the challenging $V^*$ benchmark \textbf{without retraining the vision encoder or LLM}. The example in the figure above demonstrates an instance where CLIP cannot correctly get the answer because (a) in the cropped version of the image the person in question is not present, and (b) if we use a padded image the person will be too small to provide meaningful signal to the model.}
    \label{fig:abstract}
    \vspace{-1em}
\end{figure}

\section{Introduction}
Multimodal Large Language Models (MLLMs) have shown impressive multi-modal question answering ability, yet recent work has highlighted a deficiency whereby these models struggle to answer questions about fine-grained visual details \citep{shi2024we, wu2024v}. 
MLLMs, like the popular LLaVA family \citep{liu2024improved, liu2023visual}, use a vision encoder and visual projector to embed visual information into a shared visual-linguistic embedding space before passing these tokens to a downstream LLM. This strategy is not without its flaws. Firstly, authors have observed that a high number of visual input tokens can be removed without significantly affecting performance, indicating redundancy \citep{hu2024matryoshka, li2025semantic, sun2025lvpruning}. Secondly, it has been shown that models like LLaVA overly rely on information from the vision encoder's \cls token, which captures global semantics, to answer questions \citep{zhang2024cls}. 

We posit that this failure on fine-grained VQA tasks is neither a deficiency in the training process of the MLLM nor a deficiency in the representations which can be encoded by the vision encoder. Prior works that have aimed at modifying the vision encoder or projector have implicitly assumed that the failure mode is caused by CLIP itself, but this is only partially true. \cite{li2025semantic} show that the vanilla LLaVA architecture with the CLIP encoder is capable of much better VQA performance, but requires the ``correct'' tokens to be fed to the language model. Similarly, \cite{li2024erroneous} show that the information from the CLIP encoder is often sufficient for certain vision tasks or VQA, however, the models often do not adequately use the given information. Thus, there remains room to improve MLLM performance by focusing on better use of the available tokens.

We argue that the failures incurred while using the CLIP encoder can be attributed to two specific biases induced by the inductive priors implicitly assumed during CLIP training. \textbf{Mesoscopic Bias} occurs because CLIP uses a uniform grid-patchification (UGP) strategy \citep{dosovitskiy2020image, radford2021learning} and manifests as downstream models implicitly treating uniform grid cells at a specific image scale as the fundamental unit of semantic meaning. \textbf{Interpolation Bias} arises as a consequence of CLIP being trained with fixed positional embeddings on fixed-resolution images and prevents CLIP from natively handling high-resolution images.

This work addresses the biases inherent to standard grid-patchification in MLLMs. While newer vision encoders are similarly able to handle high-resolution images, the biases we explore here remain relevant to a broad class of encoders which see continued use. Our method is specifically designed to improve the semantic content of the image tokens for the purposes of VQA.

Previous work has focused on training new vision encoders to replace CLIP \citep{guo2024llava, liu2024improved, luo2024feast, shi2024we}, but these proposals require re-training the entire MLLM, which is expensive and often not feasible. In this work, we take a minimally invasive approach and carefully reason through the consequences of updated vision priors. This leads us to a \emph{lightweight, content-aware, drop-in} modification to the CLIP encoder which we call \modelname, a portmanteau of ``quadtree'' and ``CLIP''.

\modelname empowers CLIP based MLLMs to automatically process arbitrary resolution input images, while adaptively scaling the number of input tokens based on the semantic content of the image. We find that reducing the number of input tokens has beneficial effects beyond faster computation: token reduction can decrease model hallucination and improve fine-grained VQA. To assess both the effectiveness and efficiency of \modelname, we apply our ideas to the LLaVA-1.5 family of MLLMs for VQA. Our method is particularly suited for fine-grained visual tasks like the challenging $V^*$ benchmark \citep{wu2024v}. Our method achieves a 13.6\% improvement on $V^*$, reduces hallucination rates as measured by the POPE F1 score \citep{li2023evaluating} by 5.2, and yields 
improvements across other multi-modal benchmarks including MME \citep{fu2023mme} and RealWorld-QA \citep{xai2024realworldqa}.

We accomplish this by using two novel strategies. First, to address the mesoscopic bias we introduce a non-uniform patchification scheme based on image quadtrees \citep{hunter1979operations}. Our quadtree patchification is \emph{adaptive, tunable, and training-free}, and implicitly treats semantically similar regions of the image as the fundamental unit of semantic meaning instead of UGP. Second, to address the interpolation bias, we train a small MLP network to interpolate the fixed positional CLIP embeddings while maintaining usable positional signals for downstream models. 

Our key contributions are as follows:

\begin{enumerate}[leftmargin=2em]
\item We identify two fundamental biases in the CLIP vision encoder, i.e. mesoscopic bias and interpolation bias, and propose quantitative measures of both.
\item We introduce \modelname, a lightweight, drop-in modification for CLIP that supports arbitrary image resolutions and adaptively scales the number of the image tokens based on image content. \modelname directly mitigates the aforementioned biases without retraining the vision encoder or LLM.
\item We empirically validate the effectiveness of \modelname by integrating it into the LLaVA model family \citep{liu2024improved, liu2023visual} and demonstrate substantial performance improvements. Our results are achieved without retraining the vision encoder or LLM. For the challenging $V^*$ benchmark, we achieve a significant improvement of \green{+13.6\%} accuracy using LLaVA 13B with \modelname, outperforming the previous SoTA CLIP-based LLaVA results by \green{+3.1\%} \citep{shi2024we}.
\end{enumerate}
\section{Why CLIP Fails at Higher Resolutions}\label{sec:clip}
\begin{figure}
    \centering
    \includegraphics[width=0.8\linewidth]{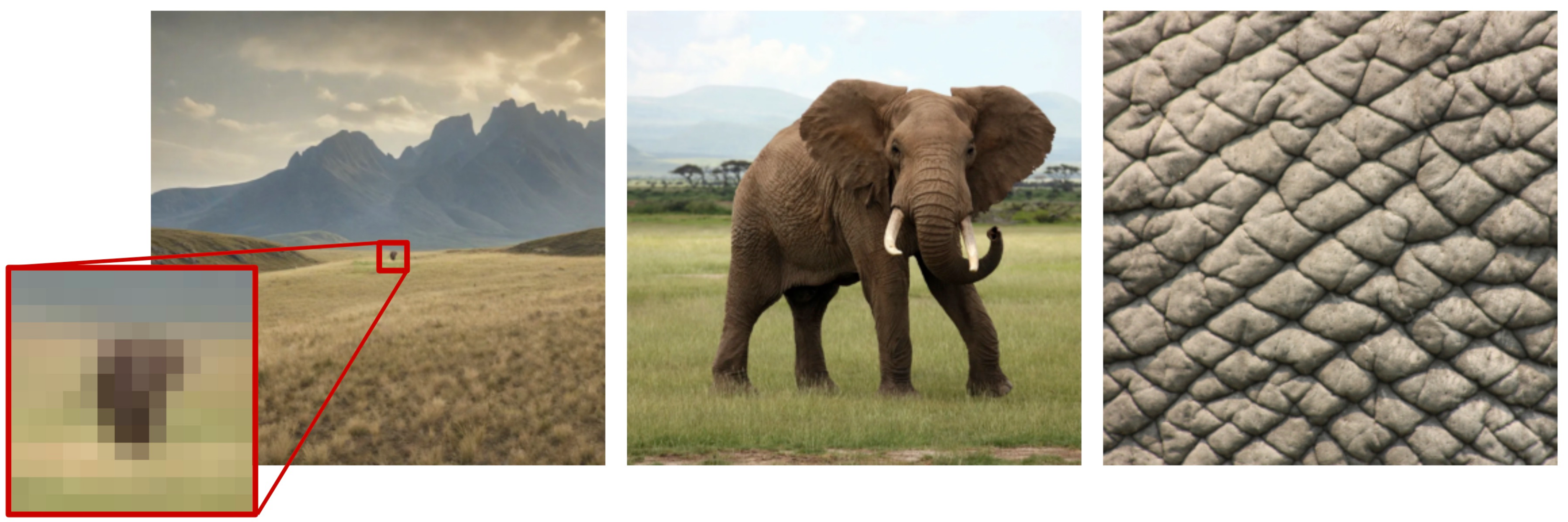}
    \vspace{-0.5em}
    \caption{An example of the same semantic feature (\texttt{animal:elephant}) at three different spatial scales. These photos could be accompanied by the question \texttt{What animal is shown in this photo?} For the leftmost image the elephant fits into a single patch. Without memorization it is unlikely for any classifier to be able to accurately identify the pixelated blob as an elephant instead of, for example, a horse or a buffalo.}
    \label{fig:macro_meso_micro}
    \vspace{-1em}
\end{figure}

The CLIP vision encoder is trained at a fixed input resolution using learned absolute positional encodings \citep{radford2021learning}. This design introduces two notable and consequential biases. First, because the positional encodings are absolute rather than relative, they do not generalize beyond the spatial grid used during training. 
Second, the encoder is trained exclusively on fixed-scale images, which biases the encoder towards only recognizing features at a specific mesoscopic spatial scale. For an exaggerated example, consider the elephants in Figure~\ref{fig:macro_meso_micro}. The CLIP encoder is most likely to understand the middle (mesoscopic) image as containing an elephant, rather than the left or right images. This is because during training it is unlikely that the leftmost image would be labeled as having an elephant in it and the rightmost image may be too zoomed in to distinguish it from other concepts. In practice the bias is not this extreme, but as we show in Figure~\ref{fig:mlp_vs_bicubic_bias} below, changing the image resolution by only a few pixels already substantially decreases the model's ability to recognize the semantic content of an image.

\paperpar{Quantification of Interpolation Bias}
Consider a single image $\mathcal{I}$ rendered at two different resolutions, $R_1 = (H_1, W_1)$ and $R_2 = (H_2, W_2)$. We denote the corresponding resized images as $\mathcal{I}_{R_1}$ and $\mathcal{I}_{R_2}$, respectively. Since both images originate from the same source and contain identical (or nearly identical) semantic content, one would reasonably expect the CLIP \texttt{[CLS]} token embedding to remain invariant or at least approximately constant across these resolutions, especially when $R_1$ and $R_2$ are only slightly different. Under this assumption, the cosine similarity between the corresponding CLIP embeddings, $\mathcal{E}_1 = \text{CLIP}(\mathcal{I}_{R_1})$ and $\mathcal{E}_2 = \text{CLIP}(\mathcal{I}_{R_2})$, serves as a measure of the deviation introduced by resolution changes. To quantify the extent to which positional embeddings contribute to this deviation, we define the interpolation bias as:
\begin{align}\label{eq:b_interp}
\mathcal{B}_{\text{Interp}}(\mathcal{I}) := \left|\left|\,\nabla_\mathcal{P}\text{CS}(\mathcal{E}_1,\mathcal{E}_2)\,\right|\right|_2 ,
\end{align}
where $\mathcal{P}$ denotes the additive positional encodings applied to patch embeddings during the CLIP encoding process \citep{radford2021learning} and CS is cosine similarity. 

\paperpar{Quantification of Mesoscopic Bias}
Mesoscopic bias is easier to quantify because we can simply remove the positional encodings and look at the cosine similarity of the \texttt{[CLS]} token embeddings at different image sizes. To this end, consider an image $\mathcal{I}$ with resolution $N\times N$ and then consider the same image rescaled to $336\times 336$, which we denote $\mathcal{I}_{336}$. Let $\mathcal{E}^z=\text{CLIP}^z(\mathcal{I})$, $\mathcal{E}_{336}^z=\text{CLIP}^z(\mathcal{I}_{336})$ be the respective \texttt{[CLS]} embeddings after setting the positional encodings to zero. Then
\begin{align*}
    C_{N\rightarrow 336}^z:=\text{CS}(\mathcal{E}^z,\mathcal{E}_{336}^z)
\end{align*}
captures the degree to which the overall embedding has changed as an effect of the mesoscopic scale of the input images.
\begin{figure}
    \centering
    \includegraphics[width=0.8\linewidth]{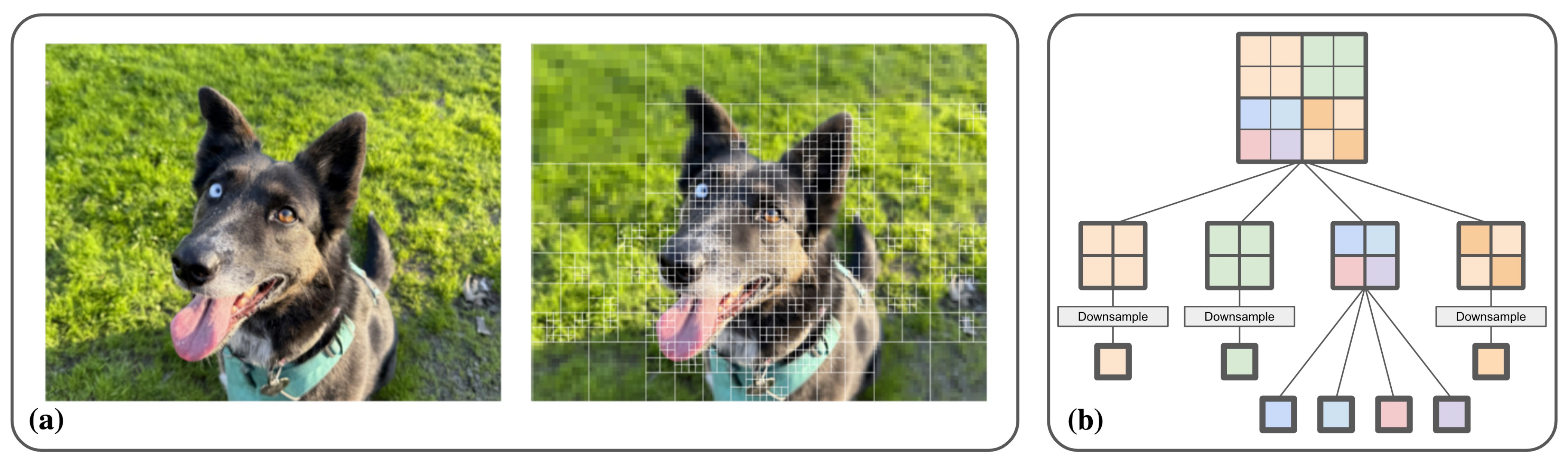}
    \vspace{-0.5em}
    \caption{\textbf{(a)} An example of the quadtree patchification (QtP) applied to a high-resolution image. QtP uses only 25\% of the original number of tokens yet retains a high-degree of semantic information. Photo courtesy of first author. \textbf{(b)} A schematic of a $4\times 4$ patch image being decomposed into $7$ leaf patches using a quadtree. Leaves which consist of more than a single patch are downsampled to the patch size.}
    \label{fig:quadtree_demo}
    \vspace{-1em}
\end{figure}

\section{Addressing the Mesoscopic and Interpolation Biases}
We identify and address two implicit inductive priors underlying the CLIP encoder, noting that these assumptions were likely adopted primarily for engineering practicality. 

The first prior is that UGP captures fundamental units of semantics. We address by replacing UGP with a \emph{content-aware quadtree patchification} (QtP). The second prior is that images can be effectively represented by center-cropping and rescaled to a fixed resolution which we address by training a small interpolation network.

\subsection{Vision Quadtree Mitigates the Effects of Mesoscopic Bias}
Natural images do not contain uniformly distributed information throughout their sub-images. In general, semantic information can continue to be extracted even when large portions of the image are subjected to extreme levels of information degradation at the pixel level (see Figure~\ref{fig:quadtree_demo}). This is why compression algorithms like JPEG work \citep{wallace1992jpeg}.

We derive a strategy for adaptively merging adjacent patches in an attempt to increase the quality of the visual signal coming from the vision encoder. This strategy is based on the intuition that many pixels in a given image do not contribute to the representation of the semantic content of the image. We propose using a quadtree \citep{hunter1979operations} structure to adaptively select tokens based on some property intrinsic to the sub-images themselves. Quadtrees, as applied in image processing, are hierarchical image representation trees which generalize a binary tree into two dimensions. At the root of the tree is the original image, and at each level we subdivide the image into four, until we reach leaf nodes which represent patches (see Figure~\ref{fig:quadtree_demo}, \textbf{(b)}). We can then prune the tree according to some selection criteria and the resulting leaf-nodes will consist of all sub-images which satisfy some maximal condition. We apply downsampling to the leaf-nodes which are larger than the CLIP encoder's patch size to obtain a sequence of patches that can be fed to the CLIP vision encoder. In theory, semantically irrelevant portions of the image are downsampled back to the mesoscopic scale that CLIP expects, and important tokens which represent a small portion of the visual field are effectively upsampled into the same scale (see Figure~\ref{fig:quadtree_demo}, \textbf{(a)}).

In what follows, we use the following quadtree selection criteria, which can be thought of as the maximum of the average gradient over a patch. Thus, an image $I$ is a leaf-node if it cannot be sub-divided or if
\begin{align}\label{eq:deriv_condition}
    \mathcal{D}(I):= \max_{x, y}(\,\partial_xI + \partial_yI\,) < \alpha,
\end{align}
where $\alpha$ is a pre-chosen selection constant.  We also test a random selection strategy as an ablation for our selection strategy. More details are contained in Appendix~\ref{app:qt_selection}.

\subsection{Coordinate-Based MLP Mitigates the Effects of Interpolation Bias}\label{sec:coordinate_mlp}
The CLIP vision encoder consists of two mechanisms that work in concert to map information from the pixel space into the embedding space. 

Let $\mathcal{P}=\{p_i\}_{i=1}^N$ be a set of patches with coordinates $\mathcal{X}=\{(x_i, y_i)\}_{i=1}^N$, $(x_i, y_i)\in [-1,1]^2$. The CLIP encoder can be understood as taking $\mathcal{P}$ and $\mathcal{X}$ and producing a sequence of tokens $\mathcal{S}=\{s_i :=\bm{E}(p_i)+\bm{M}(x_i, y_i)\}_{i=1}^N$ along with a \cls token $\bm{E}_{\cls}(\mathcal{P})$ which captures global information about the image.

CLIP is trained on $336 \times 336$ images decomposed into a series $14\times 14$ patches using the standard UGP \citep{dosovitskiy2020image, radford2021learning}. There will then be $24\times 24 = 576$ patches and to each of these patches CLIP associates a positional embedding $\mathcal{E}_{ij}\in \mathbf{R}^{1024}$, where $1\leqslant i, j\leqslant 24$ respectively index the rows and columns of both $\mathcal{E}$ and the grid of patches. For this patchification we have $\bm{M}(-1+\frac{2i}{23}, -1+\frac{2j}{23})=\mathcal{E}_{ij}$. We will extend $\bm{M}$ to the entire square $[-1, 1]^2$ so that we can natively handle images of any resolution and apply our QtP. We choose to train an MLP using our new inductive priors. Choosing an MLP for this task gives us a high-degree of expressivity.

We make the assumption that the \cls token should remain invariant when CLIP is applied to a $336\times 336$ image and the same image at its native resolution\footnote{This assumption is better justified when the native resolution is very close to $336\times 336$. However we find that generalizing this assumption to arbitrary image resolutions leads to favorable results}. Thus, if $\mathcal{G}$ is the standard UGP associated to the image $I_{336}$ and $\mathcal{P}$ is a patchification associated to the image $I_N$, then we expect that 
\begin{align}\label{eq:L_loss}
    L_{\texttt{[CLS]}}:=||\,\bm{E}_{\texttt{[cls]}}(\mathcal{G}) - \bm{E}_{\texttt{[cls]}}(\mathcal{P})\,||_{L^2} = \text{small}.
\end{align}
This provides a target for training the MLP. However, in practice $L_{\cls}$ is insufficient for training since the transformer pooling which generates the \cls token means that as long as $\sum_{ij}\mathcal{E}_{ij} = \sum_{i}\bm{M}(x_i, y_i)$, then the \cls embedding will be constant. Because we are attempting to train a drop-in modification for CLIP and because downstream MLLMs utilize the positional information from CLIP, we must ensure that the MLP positional embeddings match the CLIP positional embeddings on the standard $24\times 24$ grid. We add a residual $L^1$ error:\footnote{We found that $L^1$ loss was better than $L^2$ loss since we aim to get $\mathcal{R}$ to be smaller than $5\times 10^{-7}$. See Appendix~\ref{app:detailed_training} for more details.}
\begin{align}\label{eq:loss_residual}
    \mathcal{R}(\bm{M},\mathcal{E}):=\frac{1}{576}\sum_{i=1}^{24}\sum_{j=1}^{24}\left|\bm{M}\left(-1+\frac{2i}{23}, -1 + \frac{2j}{23}\right)-\mathcal{E}_{ij}\right|.
\end{align}
Thus we arrive at a suitable loss function for the MLP training:
\begin{align}
    \text{Loss} = L_{\texttt{[CLS]}} + \gamma \mathcal{R},
\end{align}
where $\gamma$ is a hyperparameter to balance the relative effects of the two components of the loss. Training is stable and we include additional training details in Appendix~\ref{app:detailed_training}.

\subsection{Training the Interpolation Network}\label{sec:mlp_train}
The coordinate-based MLP must be trained; however, training cost is small relative to MLLM retraining (11 hours on four NVIDIA L40S GPUs). We train the MLP for 100 epochs with the Adam optimizer \citep{kingma2014adam} on the training split of the Imagenette dataset \citep{Howard_Imagenette_2019}. This dataset is a small subset of Imagenet \citep{deng2009imagenet} with only 10 classes, and consists of about 10k images. We argue that the choice of dataset does not matter much for the MLP training because the embedding function $\bm{M}$ is independent of the image content. We train with a batch size of 14, with images kept at either their native resolution or at a resolution with the smallest edge length being $560$, whichever is smaller. We kept $\gamma = 1$. For our MLP architecture we use four hidden layers and pass the input features through a Fourier features layer \citep{tancik2020fourier} with 48 Fourier features. See Appendix~\ref{app:detailed_training} for a discussion about how we chose model hyperparameters.
\section{Experimental Results}\label{sec:experimental_results}
\begin{figure}[h]
    \centering
    \includegraphics[width=0.8\linewidth]{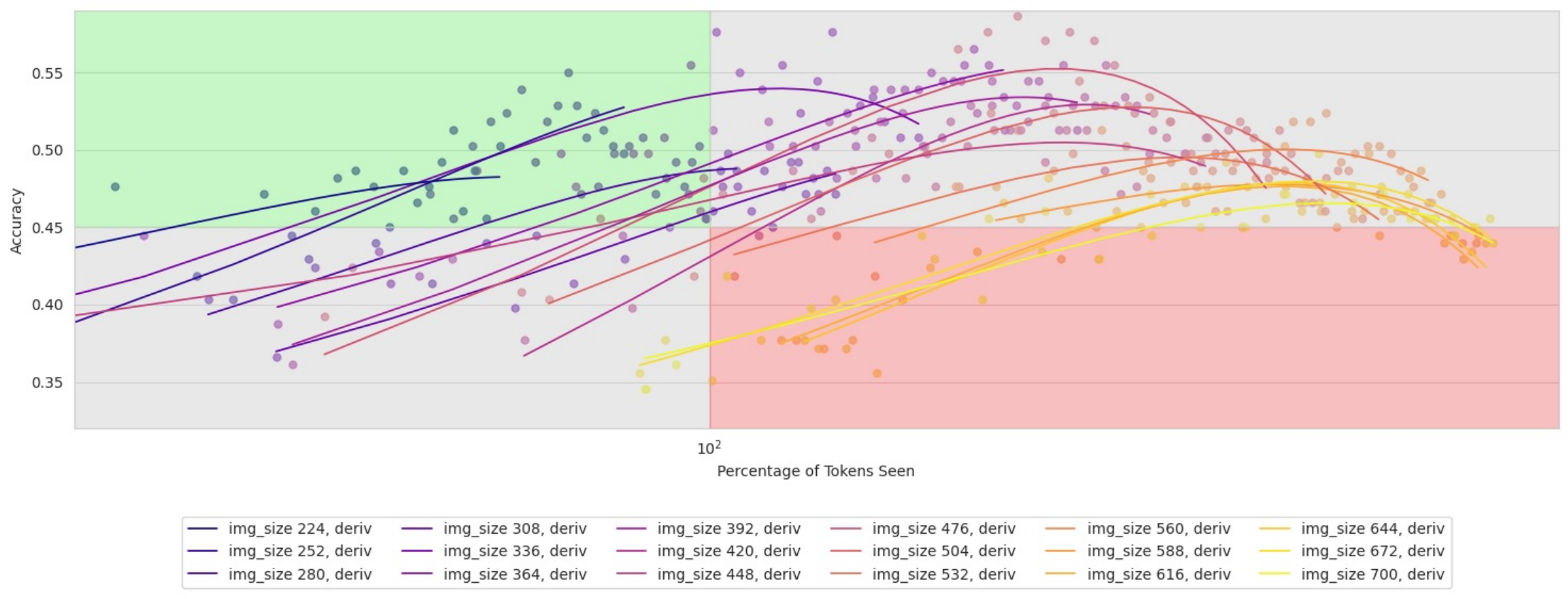}
    \vspace{-0.5em}
    \caption{The compute vs. accuracy curves for our sweep of $V^*$ with the LLaVA-\modelname-13B model. The $x$-axis is on a logarithmic scale. The green-shaded region highlights experiments where our model \textbf{surpasses the baseline} with \textbf{fewer} visual tokens.}
    \vspace{-0.5em}
    \label{fig:vstar_rainbow}
\end{figure}

Recall that the parameter $\alpha$ from equation~\ref{eq:deriv_condition} controls the amount of pruning done to the quadtree. For natural images, users can reuse the trained MLP and tune only $\alpha$; default values work reasonably well across similar domains. As Figure~\ref{fig:vstar_rainbow} illustrates, performance exhibits non-monotonic behavior with respect to $\alpha$ and image size. We recommend $\alpha > 0$ for fine-grained tasks and $\alpha \approx 0$ for general VQA. We perform sweeps in $\alpha$ and image size, over a suite of multi-modal benchmarks, in an attempt to understand the dynamics of our proposed methodology. For some benchmarks we additionally sweep native image resolution vs. cropped image resolutions. We report the best score from our sweeps in Table~\ref{tab:mmbenchmarks}. We choose to look at the performance on $V^*$ \citep{wu2024v}, MM-Bench \citep{liu2024mmbench}, POPE \citep{li2023evaluating}, CV-Bench \citep{tong2024cambrian}, the visual portion of ScienceQA \citep{lu2022learn}, MME \citep{fu2023mme}, and the RealWorld-QA benchmark \citep{xai2024realworldqa}. We use VLM~Eval \citep{duan2024vlmevalkit} to do the evaluations on MM-Bench, POPE, ScienceQA, MME, and RealWorld-QA. We use a custom evaluation script to evaluate $V^*$ and CV-Bench. More details of our experimental setup are contained in Appendix~\ref{app:detailed_eval} and instructions to reproduce our experiments are contained in Appendix~\ref{app:repro}.

\begin{figure}
    \centering
    \includegraphics[width=0.8\linewidth]{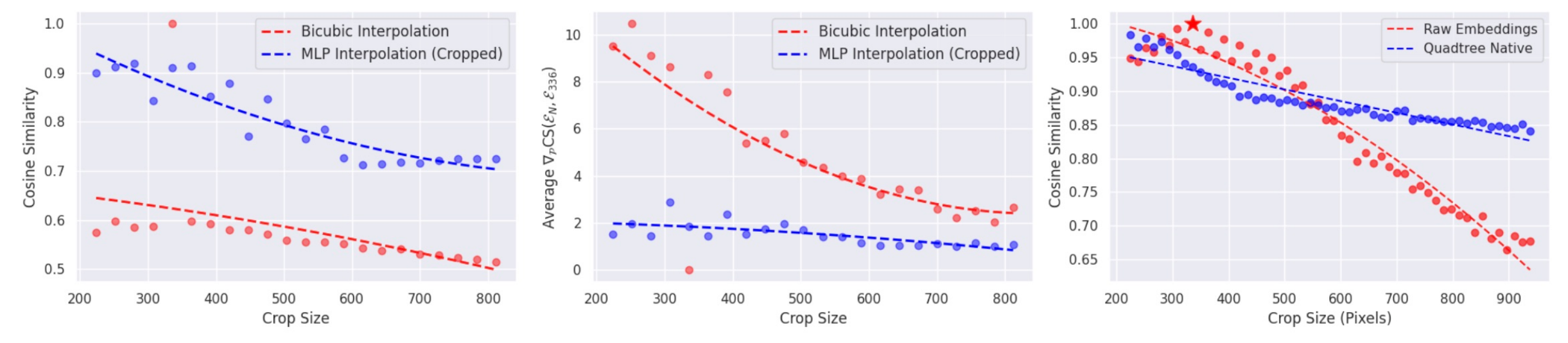}
    \vspace{-0.5em}
    \caption{The first two panels compare our MLP interpolation with bicubic interpolation. We plot $C_{N\rightarrow 336}^z$ in the first panel as a measure of mesoscopic bias and $\mathcal{B}_{\text{Interp}}$ in the middle panel as a measure of interpolation bias. The third panel shows a comparison between the \cls tokens of various image sizes with (blue) and without (red) QtP. All data is collected and averaged over the images from the $V^*$ benchmark.}
    \vspace{-0.5em}
    \label{fig:mlp_vs_bicubic_bias}
\end{figure}

\paperpar{\modelname Reduces Measured Interpolation and Mesoscopic Bias:} In Figure~\ref{fig:mlp_vs_bicubic_bias} we plot a comparison between \modelname and the vanilla CLIP encoder using bicubic interpolation, which we found outperformed bilinear interpolation. We see that MLP training successfully reduces interpolation bias as measured by $\mathcal{B}_{\text{Interp}}$ (Figure~\ref{fig:mlp_vs_bicubic_bias}, middle panel), and brings the cosine similarity between the \cls tokens together as predicted by our theoretical assumptions. Next, we observe that the quadtree selection mechanism mitigates the effects of mesoscopic bias by slowing the rate at which the cosine similarity of the CLIP \cls tokens diverge as a function of image size (Figure~\ref{fig:mlp_vs_bicubic_bias}, rightmost panel).

%
%
\begin{table*}[h!]
\centering
\caption{Performance comparison between LLaVA-QLIP and baseline LLaVA models. \textbf{Bold} highlights the better-performing variant of the same base model. \underline{Underlining} denotes the best result across all models. An asterisk (*) indicates results obtained using cropped images. Performance \green{increases} and \red{decreases} are annotated in green and red, respectively.}
\vspace{1em}
\setlength{\tabcolsep}{7pt}
{\fontsize{9pt}{13pt}\selectfont
\begin{tabular}{l|ccccccc}
\toprule[1.2pt]
Model& $V^*$ & MM-Bench & POPE F1 & CV-Bench & Sci-QA & MME & RW-QA\\
\rowcolor{codegray}
\multicolumn{8}{c}{\textit{VQA}} \\
LLaVA-1.5-7b        & 42.4              & \textbf{62.5} & 74.4              & 39.9              & \textbf{64.0} & 1207          & \textbf{49.0} \\
+ \modelname        & \textbf{53.4}     & 59.7          & \textbf{79.6}     & \textbf{40.2}     & 63.5          & \textbf{1241} & 47.3 \\
                    & \green{(+11.0)}   & \red{(-2.8)}  & \green{(+5.2)}    & \green{(+0.3)}    & \red{(-0.5)}  & \green{(+34)} & \red{(-1.7)} \\

\hline
LLaVA-1.5-13B       & 45.0                      & 67.4      & 82.4 & \underline{\textbf{61.6}} & 67.8 & \underline{\textbf{1390}} & 48.0 \\
+ \modelname        & \underline{\textbf{58.6}} & \underline{\textbf{67.9}}* & \underline{\textbf{83.6}} & 60.7* & \underline{\textbf{67.9}} & 1388*  & \underline{\textbf{49.4}} \\

                    & \green{(+13.6)} & \green{(+0.5)} & \green{(+1.2)} & \red{(-0.9)} & \green{(+0.1)} & \red{(-2)} & \green{(+1.4)} \\
\bottomrule[1.2pt]
\end{tabular}
}

\label{tab:mmbenchmarks}
\end{table*}

\paperpar{\modelname Significantly Improves the Detailed Visual Grounding on High-Resolution Images:} The $V^*$ benchmark \citep{wu2024v} is a challenging, vision centric benchmark focused on fine-grained image understanding. This benchmark is particularly challenging for CLIP-based vision encoders because the questions are designed to be answered with access to the full-resolution image (see Figure~\ref{fig:abstract}). Without access to all of the appropriate visual information the model is often reduced to guessing.

\begin{figure}[h]
    \centering
    \includegraphics[width=0.8\linewidth]{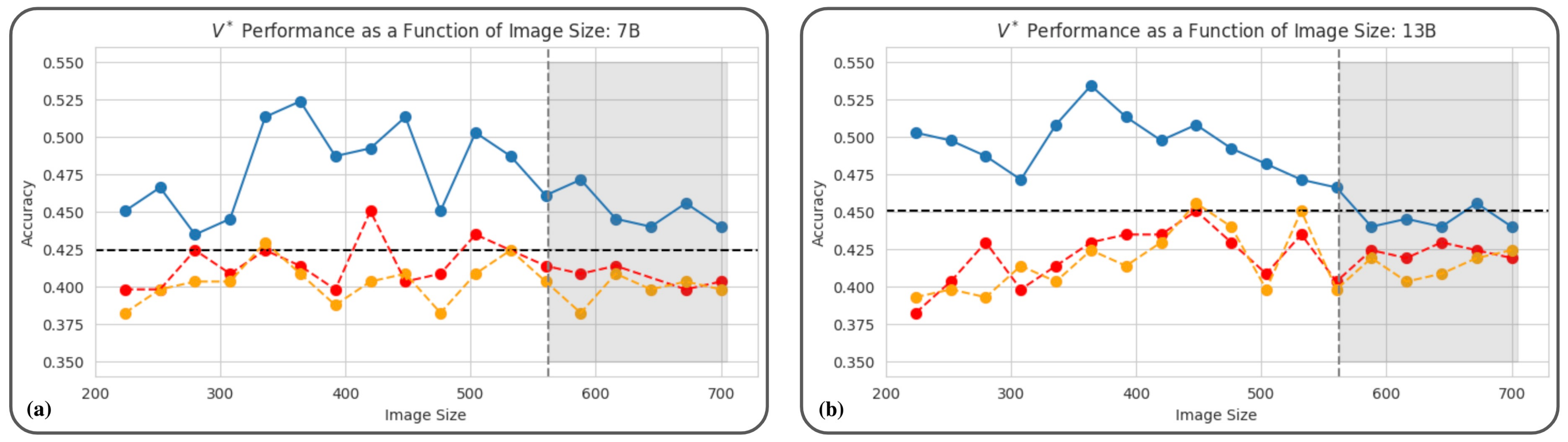}
    \vspace{-0.5em}
    \caption{The performance on $V^*$ using re-scaled and cropped images with no quadtree selection mechanism and our MLP interpolation. The red line is with bicubic interpolation and the orange line is with bilinear interpolation. The black line represents performance of the base CLIP model with $336\times 336$ cropping. The 7B model is plotted on the left, and the 13B model on the right. We see that neither bilinear nor bicubic interpolation is suitable for extending CLIP to larger resolutions.}
    \vspace{-0.5em}
    \label{fig:vstar-image-size}
\end{figure}

Figure~\ref{fig:vstar-image-size} demonstrates that in the absence of the quadtree selection method, our MLP interpolation network already allows the model to effectively utilize all of the image tokens from the original image. We note that the 7B parameter is more robust to image sizes which were not seen during training than the 13B model. These results already indicate that there is a large performance gap that can be closed with minimal interventions, indicating that a significant portion of the poor performance on high-resolution image tasks can be explained simply by a lack of access to high-quality visual input signal (c.f. \citep{li2025semantic}). This result indicates that the CLIP encoder and LLaVA weights possess sufficient capacity to do VQA, but lack high-quality inputs.

Figure~\ref{fig:vstar_rainbow} shows the full sweep over image size and $\alpha$, plotted with cubic best-fit lines. The $x$-axis is measured in percentage of tokens seen compared to the baseline model, on a logarithmic scale. 
We see a clear trend where increasing $\alpha$ increases performance with maximal performance occurring for $\alpha >0$. This indicates that the QtP mechanism is complementing the MLP interpolation to boost VQA performance, either by reducing the number of image tokens and sending stronger attention signal to the LLM \citep{levy2024same, velivckovic2024softmax}, by reducing noise by combining redundant image patches through merging, or both. Our ablations in Section~\ref{sec:ablations} below suggest that the latter is more likely.

\paperpar{Improved Token Efficiency:} Previous studies which reduce token counts have aimed at matching MLLM performance with fewer tokens \citep{cao2023pumer, chen2024image, hu2024matryoshka, li2025semantic, tang2022quadtree}. Our work is largely orthogonal to the aforementioned works, however we note in Figure~\ref{fig:vstar_rainbow} that we can achieve higher than baseline accuracy with \emph{fewer} image tokens than the baseline model. This is shown in the figure by the top left region, shaded in green, which represents higher than baseline accuracy with lower than baseline numbers of tokens. This reveals that the quadtree selection method prunes tokens in such a way that higher-quality visual signal is provided to the LLM. The work \citep{li2025semantic} demonstrated that such improved performance with reduced tokens is theoretically possible, but to our knowledge this work is the first time such a result has been achieved in practice.

\paperpar{\modelname Matches or Improves Performance Across a Range of MLLM Benchmarks:}
Because our method is trained to be both minimally invasive and does not require re-training of the MLLM, we can adjust the model parameters to fit the task at hand without re-training. Because our training program was oriented towards matching CLIP outputs on images which are the same size as CLIP was trained on, we can nearly achieve baseline performance for any benchmark by using $336\times 336$ images with $\alpha=0$. Any loss in performance beyond that can be attributed to the error in interpolating the CLIP embeddings with our MLP network. Notably we find little to no change in performance on MM-Bench, CV-Bench, Sci-QA, MME, or RealWorld QA. 

\modelname is designed to perform well on images with fine-grained, high-resolution semantic detail. Because of this global or coarse cues useful for general VQA may be attenuated. For example, when tuned for fine-grained performance on $V^*$, the 7B model shows regressions on MM-Bench and RealWorld-QA (Table~\ref{tab:mmbenchmarks}). The method is intentionally specialized for fine-grained tasks and reveals latent capacity in existing models rather than aiming to replace full retraining approaches.

\paperpar{LLaVA 13B is Sensitive to Image Aspect Ratio:} On three of the seven benchmarks the 13B model attained its best performance when the input images were cropped to be square at the original image resolution of $336 \times 336$ with $\alpha < 0.1$. We found that performance quickly dropped off for these three benchmarks when we varied image size or increased $\alpha$. We suspect that the 13B parameter version of LLaVA is much more sensitive to deviations in the \cls token, and the drop-off in performance seems correlated with the change in cosine similarity of the \cls token plotted in Figure~\ref{fig:mlp_vs_bicubic_bias}. We did not observe the same trend in the 7B model, nor did we observe this trend on $V^*$, where the content of the \cls token is not helpful for answering the questions.

\begin{table*}[h!]
\caption{Comparison of LLaVA-QLIP with other models which improve fine-detail grounding. We report the numbers from the authors' papers. Note that $S^2$ requires pre-training and instruction tuning of the LLM \cite{shi2024we}, and that SEAL requires fully re-placing the vision encoder before pre-training and instruction tuning \cite{wu2024v}.}
\centering
\setlength{\tabcolsep}{10pt}
{\fontsize{9pt}{13pt}\selectfont
\begin{tabular}{l|cccc}
\toprule[1.2pt]
Model   & $V^*$-Att     & $V^*$-Rel     & $V^*$ Overall     & POPE F1 \\
\rowcolor{codegray}
\multicolumn{5}{c}{\textit{Fine-grained grounding}} \\
\modelname-7B               & 50.4  & 60.5  & 53.4  & 79.6 \\
$S^2$-7B \cite{shi2024we}   & 51.3  & 61.8  & 55.5  & - \\
\midrule
\modelname-13B              & 53.9  & 65.8  & 58.6  & \textbf{83.6} \\
$S^2$-13B \cite{shi2024we}  & 50.4  & 63.2  & 55.5  & - \\
\midrule
SEAL (7B) \cite{wu2024v}    & \textbf{74.8} & \textbf{76.3} & \textbf{75.4} & 82.4 \\

\bottomrule[1.2pt]
\end{tabular}}
    \label{tab:compare}
\end{table*}

\paperpar{Hallucination can be Mitigated by Reducing the Number of Image Tokens:}
The POPE dataset was designed to measure the hallucination proclivity of MLLMs \citep{li2023evaluating}. The proposed measurement of model performance for POPE is the F1 score. For both the 7B and 13B \modelname models we saw increased performance on POPE, with more significant gains for the 7B model. In fact, \modelname even outperforms SEAL \citep{wu2024v} which is a heavily optimized version of LLaVA designed specifically to address fine-grained VQA (see Table~\ref{tab:compare}). We found that peak POPE performance occurred with the smallest image size we tested (shortest edge is $224$ pixels), and an $\alpha=0.7$, corresponding to slightly less than 50\% of the baseline image tokens. 
\section{Ablations}\label{sec:ablations}
\begin{figure}
  \centering
  \includegraphics[width=0.8\linewidth]{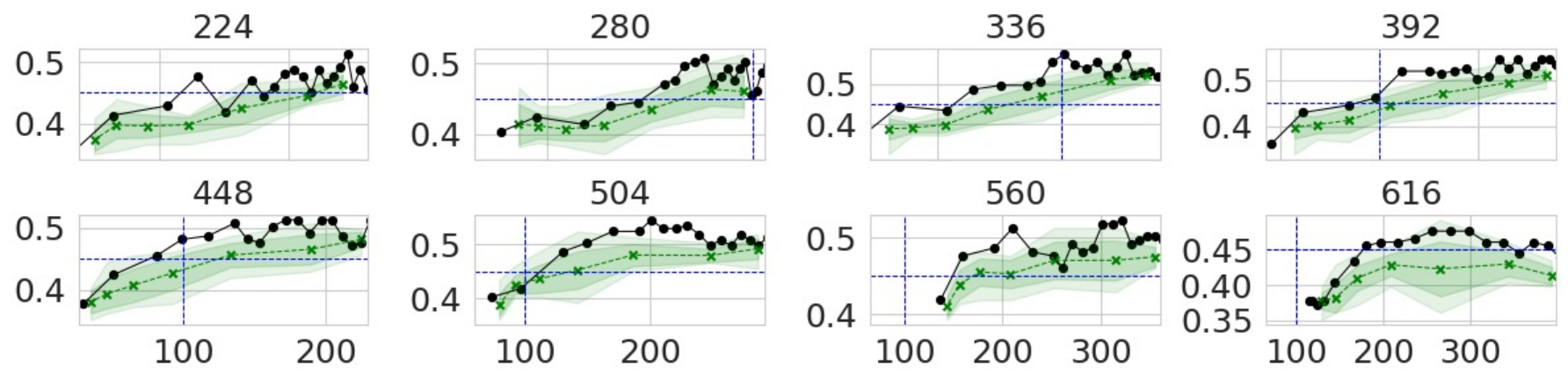}
  \vspace{-0.5em}
  \caption{Ablation on $V^*$ with \modelname-13B. The black curves are \modelname, with derivative pruning, and the green curves are \modelname with random pruning. The green curves are plotted with min/max lightly shaded, and the first standard deviation more darkly shaded. Each of the evaluations with the random selection strategy was run 10 times to compute the average and standard deviation. The $x$-axis is the percentage of image tokens seen compared to baseline, and the $y$-axis is accuracy. Each pane is labeled with its image size and the vertical and horizontal blue dashed lines represent baseline number of image tokens and baseline accuracy respectively.}
  \vspace{-1.5em}
  \label{fig:vstar_ablation}
\end{figure}

We ablate our design decisions along two axes. The first axis is along interpolation strategy, where we show that our MLP network vastly outperforms  bilinear and bicubic interpolation. Next, we demonstrate that our performance improvements from the quadtree mechanism are predicated on selection strategy and not due solely to a reduced token counts. More detailed ablations are contained in Appendix~\ref{app:ablations}.

\paperpar{MLP Interpolation is Essential for Generalizing to Arbitrary Image Sizes:}
We experiment with using bicubic interpolation to scale evaluation with image size. We find that across all of our benchmarks bicubic and bilinear interpolation under-perform our MLP interpolation. This is clearly demonstrated for $V^*$ benchmark in Figure~\ref{fig:vstar-image-size}, where the bicubic and bilinear interpolation schemes under-perform even the baseline model performance on average.

\paperpar{Performance Gains are not Solely a Result of a Reduced Number of Image Tokens:}
We verify that the derivative selection strategy provides a meaningful information signal to the downstream LLM by comparing it to using a random selection strategy which prunes quadtree branches at some random rate. We compare the performance of these two selection strategies on the $V^*$ benchmark in Figure~\ref{fig:vstar_ablation}, where keeping precise semantic information about particular regions of the image is critically important. We find that on average there are large gaps in performance between random selection and derivative selection, indicating that our derivative selection strategy provides a more meaningful visual signal to the model.
\section{Related Work}\label{sec:related}

\paperpar{Improved Vision Encoders and MLLMs}
The observation that grid patchification at a fixed image resolution is a poor inductive bias is not new. This has led to a litany of proposed replacements for CLIP \citep{radford2021learning} and ViT \citep{dosovitskiy2020image}. For example, the studies \citep{bolya2022token, darcet2023vision, dehghani2023patch, duggal2024adaptive, fan2021multiscale, haurum2023tokens, kong2022spvit, lee2022mpvit, marin2023token, meng2022adavit, oquab2023dinov2, yang2022visual, zhang2022vsa} propose modifications to the ViT architecture which provide better visual signal. These studies do not attempt to train an attendant LLM to create an MLLM. The studies \citep{bigverdi2024perception, guo2024llava, liu2024improved, lu2022learn, luo2024feast, shi2024we, thapa2024dragonfly, tong2024cambrian, wang2024adaptvision, wu2024towards, yang2022visual, zhang2025dockylin} introduce new vision encoders specifically in the context of MLLM, but require pre-training and instruction tuning. The most closely related result work to ours is by Shi et al. \citep{shi2024we} who show that LLaVA performance can be increased substantially by feeding the LLM visual tokens from different scales while keeping the CLIP encoder frozen. We go beyond all of these studies by obtaining improved performance using the \emph{same} underlying MLLM backbone, with no pre-training, instruction-tuning, or supervised fine-tuning of the language model.

Newer vision encoders such as InternVL or Qwen-V employ integrated token reduction, multi-resolution training, or relative/rotary position encodings, which address interpolation and resolution handling in a different way than we have here. These architectural choices make direct integration of \modelname non-trivial, and we leave adaptation to such encoders as future work.

\paperpar{Token Pruning and Merging}
Many MLLM studies have been directed at reducing the number of visual input tokens, either by pruning tokens or merging them. Such reductions are well-motivated. \citep{levy2024same, velivckovic2024softmax} show that in addition to being computationally expensive, feeding an LLM too many tokens can harm performance. Recent work has also demonstrated that MLLMs rely heavily on the \cls token during VQA \citep{zhang2024cls}, which helps explain why previous authors have been able to remove up to 95\% of the visual tokens and nearly maintain MLLM performance \citep{cao2023pumer, chen2024image, hu2024matryoshka, sun2025lvpruning, tang2022quadtree}, or prune tokens across video frames while maintaining performance \citep{choudhury2024don}. A key distinction is that \modelname operates \emph{before} encoding, at the image patchification stage, whereas most pruning and merging methods operate \emph{after} encoding and typically trade accuracy for efficiency. \modelname instead improves the signal quality during patchification rather than pruning encoded tokens. Additionally, all of these studies require an expensive pre-training and fine-tuning stage to align the LLM with the vision encoder. Furthermore, our work is orthogonal to the studies \citep{cao2023pumer, hu2024matryoshka, sun2025lvpruning, tang2022quadtree} since these models rely on training LLaVA family models while using the CLIP encoder, which can be replaced in their studies by \modelname.

\section{Conclusion}

We have proposed \modelname, a drop-in, adaptive, and content-aware replacement for the CLIP encoder. We defined mesoscopic bias and interpolation bias, argued that these biases are responsible for performance difficulties on fine-grained VQA, and shown that \modelname satisfactorily addresses these biases. We achieve +13.6\% accuracy on the challenging $V^*$ benchmark without retraining the vision encoder or LLM. We are also able to exceed baseline performance on $V^*$ while using fewer image tokens. On other benchmarks, we show that we can nearly match or exceed baseline performance.

\section*{Acknowledgements}
KRC was partially supported by a Lambda Labs Research Grant. BL and MC were supported by the National Science Foundation grants OAC-2531126, ITE-2333736 and an Amazon Nova Trusted AI Prize.

\bibliography{bib}
\bibliographystyle{iclr2026_conference}

\newpage
\appendix
\section{Limitations}
\paperpar{Scope and Applicability}
\modelname is not designed for dense prediction tasks such as segmentation, object detection, or visual grounding. It does not guarantee preservation of pixel-level spatial fidelity and is intended for semantic aggregation and token budgeting rather than dense spatial reasoning. We have not validated \modelname on non-natural image domains (e.g., medical imaging, satellite imagery). Integration with newer MLLMs that employ integrated token reduction, multi-resolution training, or relative position encodings (e.g., InternVL, Qwen-VL, SigLIP-style encoders) requires architectural changes and is left as future work.

\paperpar{Technical Assumptions}
In Section~\ref{sec:clip} we assumed that the \cls token should be constant as a function of image size. This assumption, while stronger than the original implicit prior of CLIP, still lacks theoretical justification. It is easy to argue that a strong vision prior could be stated. The reason for this is that the CLIP encoder's understanding of an image may change as we scale image size, despite the theoretical alignment of the semantic content. For example, in the leftmost panel of Figure~\ref{fig:macro_meso_micro} we would not expect an image in which the elephant occupies $576$ patches to have the same \cls embedding as the zoomed out version.

We did not fully sweep or optimize the MLP training due to compute limitations (c.f. Appendix~\ref{app:detailed_training} for a discussion of how we arrived at our hyper-parameters). Sweeping the MLP training hyper-parameters more fully would likely yield a better MLP model.

We trained the MLP on images that were smaller than or equal to $560$ on their shortest edge. This was primarily an exercise in the tradeoff between batch-size and image-size. Training on larger images is preferable, but at the cost of smaller batch-sizes and significantly longer training times. We found that $560$ was a happy medium for this trade-off. Future work could explore ways to train the MLP on very large images without actually loading the entirety of the image into memory. We believe such a methodology would be useful more broadly in the computer vision / multi-modal communities.

We also did not explore training the MLP on different datasets. While we believe that the content of the images is largely immaterial we suspect that the distribution of image sizes is quite important. We leave an investigation of this relationship to future work.

While we did explore multiple selection strategies (c.f. Appendix~\ref{app:qt_selection}), there is room for a more comprehensive and theoretically justified exploration of potential selection strategies. For example one could run a large-scale study correlating different selection methods with how well they find the "correct" tokens predicted by \citep{li2025semantic}.

Finally, we could run our model through a more comprehensive suite of benchmarks to gain a more accurate sense of its performance.

\section{Implementation Details}
\paperpar{Inference Overhead}
Quadtree construction overhead is small compared to the CLIP forward pass. Inference time is dominated by token count. The procedure can be optimized substantially in compiled implementations, and with KV caching, quadtree construction is amortized.

\paperpar{Quadtree Construction}
To build a quadtree out of patches requires an image to be (a) square with (b) sidelengths consisting of $2^N$ patches for $N\in \mathbb{N}$. Obviously we are able to apply our methodology to images which are not of this size and we explain how we do so.

For concreteness, suppose we are given an image consisting of $M\times N$ patches. We can always center crop the images to the nearest patch size at a loss of at most 13 pixels. For this paper we first resize the smallest edge to our target size, then center-crop the image so that the longest side is also an integer number of patches.

\begin{figure}[b!]
    \centering
    \includegraphics[width=0.92\linewidth]{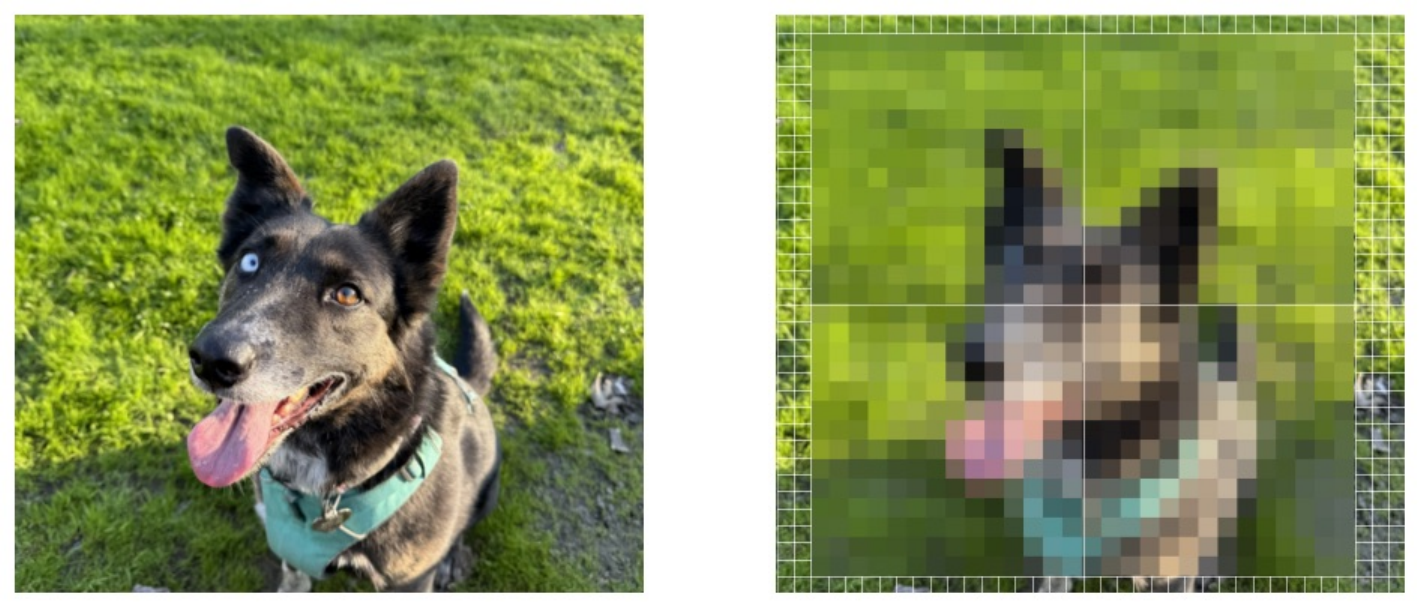}
    \caption{A quadtree applied to the image used in Figure~\ref{fig:quadtree_demo}, except with a different image size. The image in Figure~\ref{fig:quadtree_demo} is $672\times 896$, which can be decomposed into a $3\times 4$ grid of $224\times 224$ sub-images. Since $224=2^4\times 14$ each of these 12 sub-images can have a QtP. The image in this figure is $476\times 518$, which cannot be divided into QtP sub-images. The maximal grid of QtP-enabled sub-images is the $2\times 2$ grid of $224\times 224$ sub-images which are outlined in this Figure. Note the remaining patches are left as is around the border of the image.}
    \label{fig:qtp_example}
\end{figure}

Next, we find a grid of sub-images which maximally covers the original image, and where each of the sub-images in the grid is square with side lengths of $2^P$ for some $P$. For the remaining patches we leave them as is and pass their embeddings to the LLM. This process maximizes the number of patches in the image that can be subject to QtP. See Figure~\ref{fig:qtp_example} for an example of this methodology applied to an image.

For natural images, users can reuse the trained MLP and tune only $\alpha$; tuning $\alpha$ is recommended when optimizing for a specific benchmark or resolution. For full implementation details see our code at \url{https://github.com/KyroChi/qlip} (and Appendix~\ref{app:repro}).

\section{Detailed MLP Training}\label{app:detailed_training}
\subsection{Hyperparameters and Training Setup}
In our ad-hoc testing we quickly determined that for benchmark performance the MLP interpolation error was much more significant than the overall \cls embedding error. Therefore, our subsequent training experiments were targeted primarily at reducing MLP error.

We did not perform an extensive hyperparameter sweep over MLP architectures because the cost was prohibitive given our available compute. In what follows we describe our findings as we manually swept in individual directions to ablate our training hyperparameters.
\begin{itemize}
    \item $L^2$ loss for \cls tokens is better than cosine similarity.
    \item $L^1$ loss for interpolation loss is better than $L^2$ loss. We found that the $L^2$ version of \eqref{eq:loss_residual} was made smaller during training if we used the $L^1$ loss as the actual training target. We suspect that this is because the $L^2$ loss gets quite small (on the order of $10^{-5}$ to $10^{-7}$, and stops sending meaningful signal to the weights.
    \item We swept four orders of magnitude for $\gamma$, $\gamma=10^3, 10^2, 10, 1, 0.75$. We found that $\gamma = 1$ produced the best results and had the most stable training dynamics.
    \item Training on larger images produces better results but is more computationally expensive. The larger images seemed to give better results but took too long to perform meaningful sweeps over. We opted for a small batch size of 14 since it accelerated training while continuing to produce satisfactory results. We arrived at this number by choosing the maximal image size we were willing to train on and then saturating the GPUs.
    \item Dynamics appear stable regardless of batch size. We found that even with a very small batch size of 1 or 2 the training remained stable.
    \item Depth 4 MLP is better than a depth 2 MLP. We found that increasing the MLP depth from 2 to 4 gave better results and faster convergence of the interpolation error. We did not try depths greater than 4.
    \item Fourier features: We tried 16, 32, and 48 Fourier features and found that 48 yielded the best results.
    \item We used a cosine learning rate scheduler and did not experiment with adjusting the schedule. See our code for full details.
    \item Learning rate. We experimented with various learning rates and found that $7.5\times 10^{-5}$ was a good learning rate. We experimented with higher learning rates and found that they made the training unstable.
    \item We use a hidden width of 1024 and did not experiment with other widths.
\end{itemize}

During training we use a learning rate of $7.5\times 10^{-5}$ with the Adam optimizer using the default PyTorch configuration. Our MLP has four hidden layers and 48 Fourier features. We train for 100 epochs using a standard cosine learning rate scheduler. During training we use a quadtree with a $10\%$ random merging strategy. The reasoning for this is two-fold. First, introducing the random merging allows the model to see more patch locations during training, and thus effectively increases the image sizes that our model can handle. Second, it acts as a regularizer to prevent overfitting to the data-distribution of our training dataset. This is because with a deterministic sampler the positional encodings would align themselves to common QtP patterns. For example, if the objects in the training data were centered and the background had low semantic content, the model may overfit to such a situation and not be robust to situations in which the objects of interest are not centered in the image.

\subsection{Final Training Curves}
\begin{figure}
    \centering
    \includegraphics[width=0.92\linewidth]{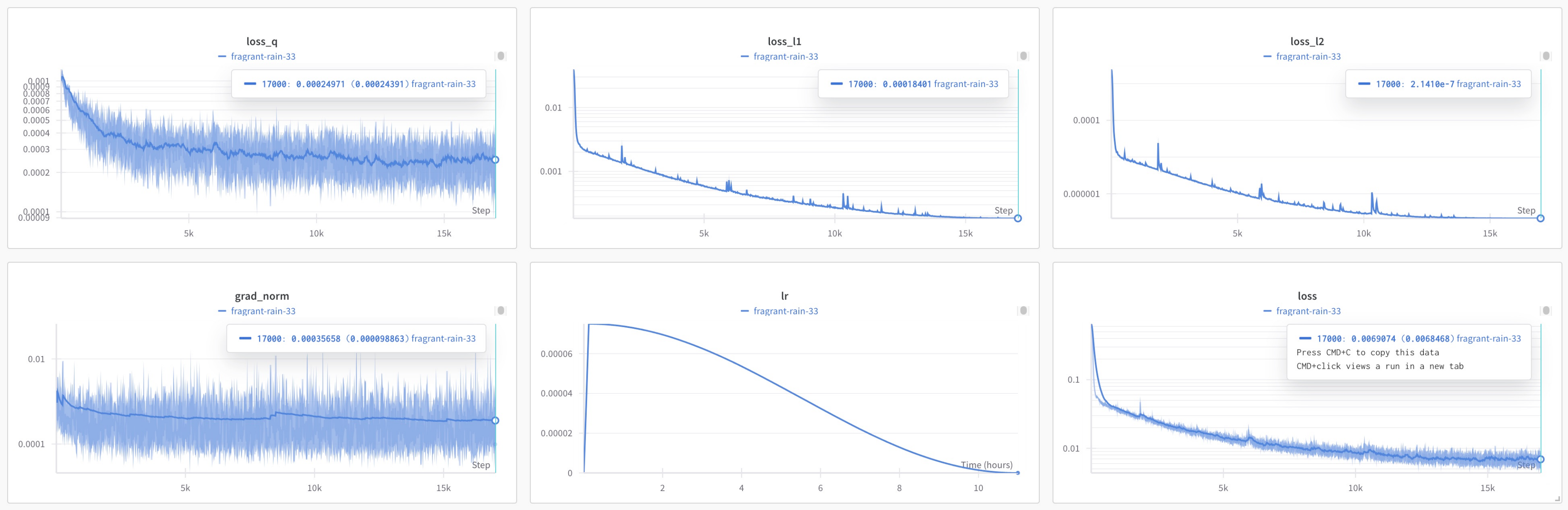}
    \caption{The training curves for our MLP training run. In the upper left is the $L_{\cls}$. The upper middle is the residual loss $\mathcal{R}$ from equation~\ref{eq:loss_residual}. The upper right is the $L^2$ analog of $\mathcal{R}$. Bottom left is the grad norm with respect to the positional encodings, given by $\mathcal{B}_{\text{Interp}}$ in equation~\ref{eq:b_interp} above. The bottom middle is our learning rate as a function of time. The bottom right is the training loss, which is the sum of the upper left and upper middle panels. We did not stop training when spikes occurred and we found that the loss spikes were transient. The upper row is plotted with a logarithmic $y$-axis, as are the bottom left and bottom right panels.}
    \label{fig:training_curves}
\end{figure}

The noise in the $L_{\cls}$ term and the grad norm terms is expected as a consequence of training on multiple resolutions simultaneously, as well as using the random selection strategy during training. We observed that the variance of these curves decreases if we restrict training to a narrower band of resolutions and / or remove the random selection from the training. The full training curve appears in Figure~\ref{fig:training_curves}

\subsection{Disclosure of Additional Computing Resources}\label{app:training_disclosure}
We report that training the MLP took 11 hours in Section~\ref{sec:mlp_train}. This time does not account for the hyperparameter sweeping that we did, nor does it account for the experimentation and development phase of our methodology. We did not keep track of the GPU hours that were used during the completion of this project. We had access to two 4x RTX 6000 machines, one 4x L40s machine, and one 8x RTX 6000 machine. We variously used compute on these three machines as it became available. Machines are shared between the members of our research group.

\section{Extended Experimental Results}
\subsection{More Details on Evaluation Strategies}\label{app:detailed_eval}
We chose parameter sweep ranges through ad-hoc probing of both image size and $\alpha$ values. Once we found good enough endpoints we would sweep the values in-between, keeping the cost of evaluations in mind as we chose our sweep parameters. We would stop sweeping early if the results were trending in the wrong direction, since over regularization from QtP is expected to cause consistent declines in performance after a certain threshold.

\subsection{Native Image Resolution vs. Cropped}
For models that did not perform well-enough using the native resolution we would switch to sweeping the cropped versions. In one case we report numbers from our model with no QtP and only the MLP interpolation (MME).

\subsection{MME Benchmark}
We found that performance with native images was poor on MME so we swept cropped images. For the 7B model this lead to overperformance of the baseline, but for the 13B model we could not get overperformance of the baseline with $\alpha > 0$.  Our top performing 13B model was with $336\times 336$ image size, random selection, and $\alpha=0$. This represents our model's closest approximation to the baseline model and any error is accounted for by the numerical error in the bicubic interpolation procedure. MME evaluations are expensive.

\begin{figure}
    \centering
    \includegraphics[width=0.92\linewidth]{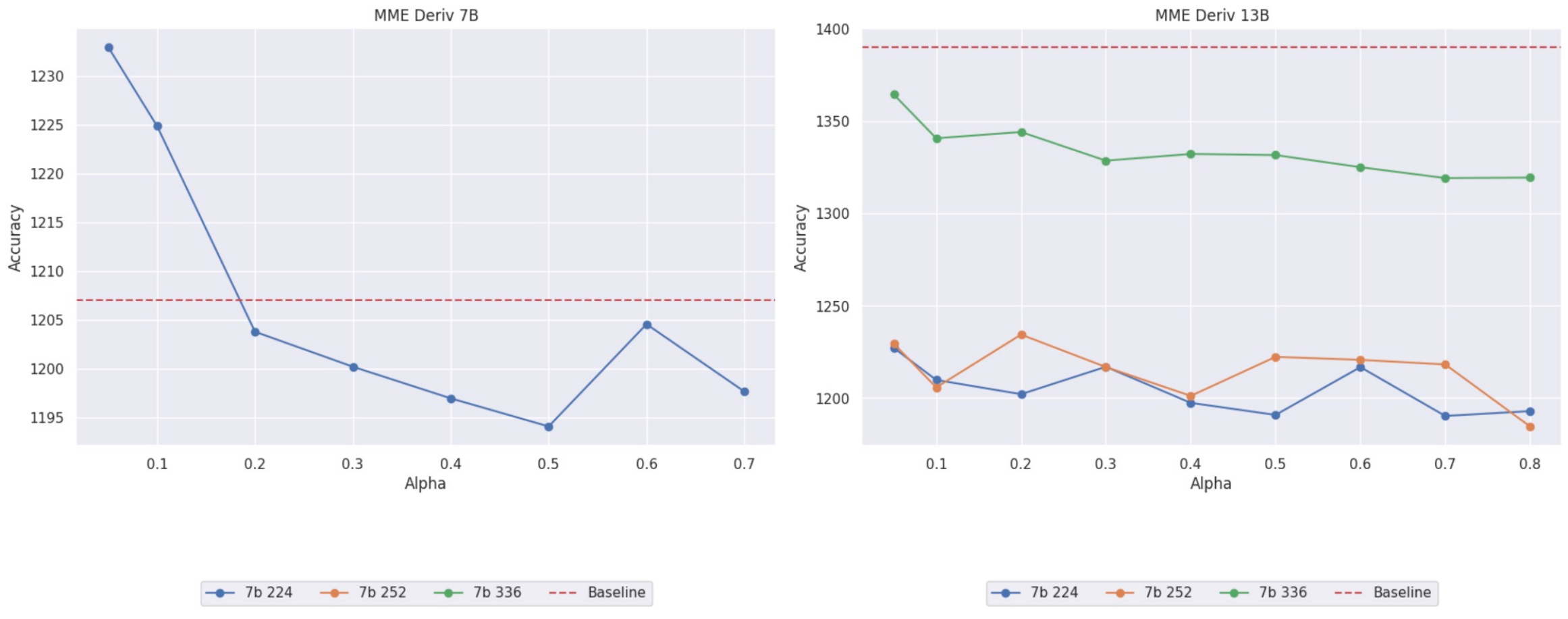}
    \caption{Sweeps on the MME benchmark \citep{fu2023mme}. Baselines are indicated by the dashed red line.}
    \label{fig:enter-label}
\end{figure}

\subsection{MM-Bench}
\begin{figure}
    \centering
    \includegraphics[width=0.92\linewidth]{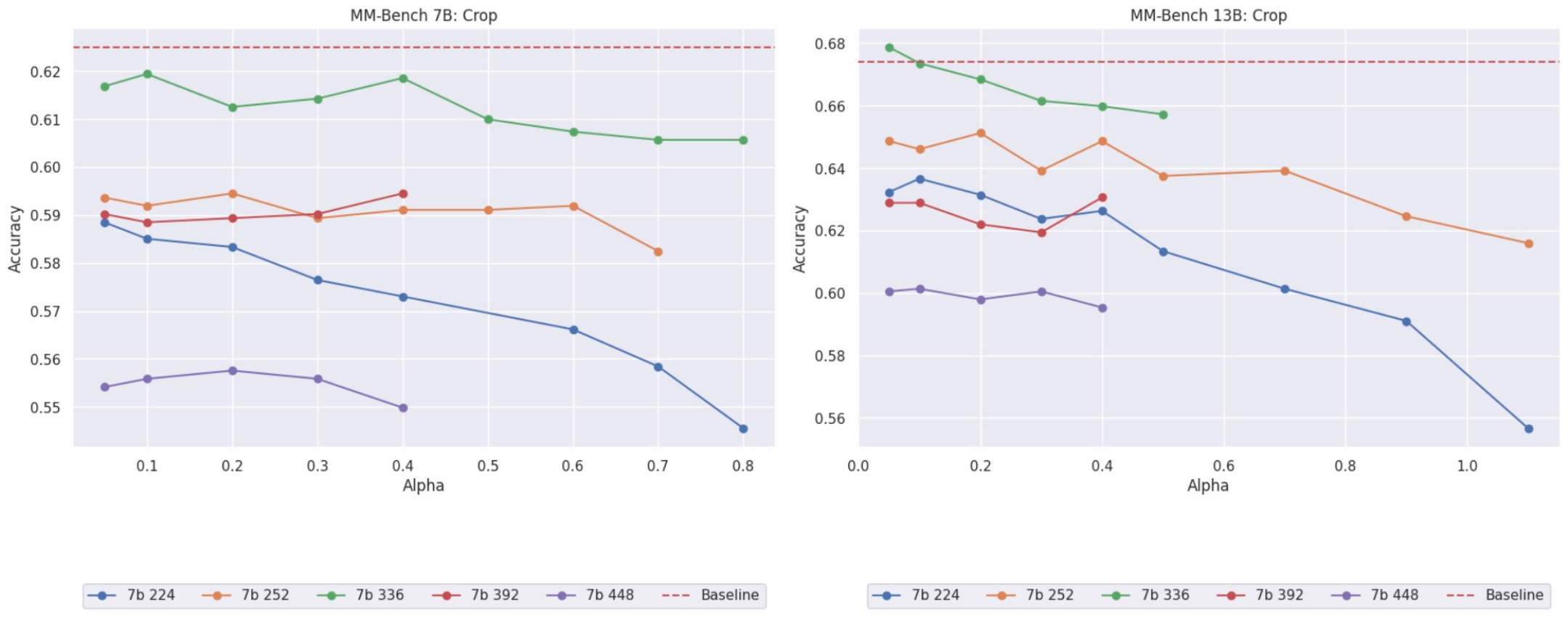}
    \caption{Sweeps on the MMBench benchmark \citep{liu2024mmbench}. Baselines are indicated by the dashed red line.}
    \label{fig:mmbench_sweep}
\end{figure}

For MM-Bench we found that the results were better with cropping than using the native image resolution. We swept several image sizes, but pruned the sweeps if the performance was proving poor. We swept image sizes of $224, 252, 336, 392$ and $448$. The results of our sweeps, including the specific $\alpha$ values chosen for each image size are shown in Figure~\ref{fig:mmbench_sweep}.

\subsection{SciQA}
\begin{figure}
    \centering
    \includegraphics[width=0.92\linewidth]{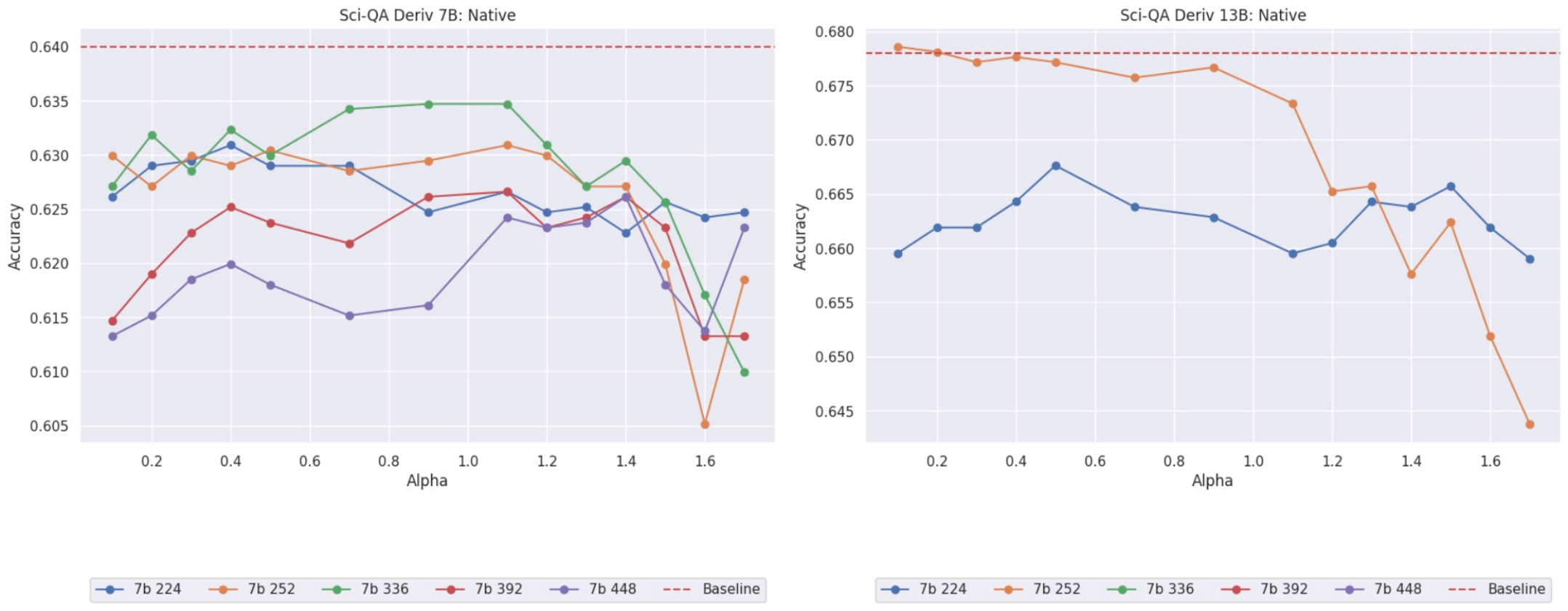}
    \caption{Sweeps on the ScienceQA benchmark \citep{lu2022learn}. Baselines are indicated by the dashed red line.}
    \label{fig:sciqa_sweep}
\end{figure}

See Figure~\ref{fig:sciqa_sweep}.
Figure~\ref{fig:sciqa_sweep}

\subsection{POPE}
\begin{figure}
    \centering
    \includegraphics[width=0.92\linewidth]{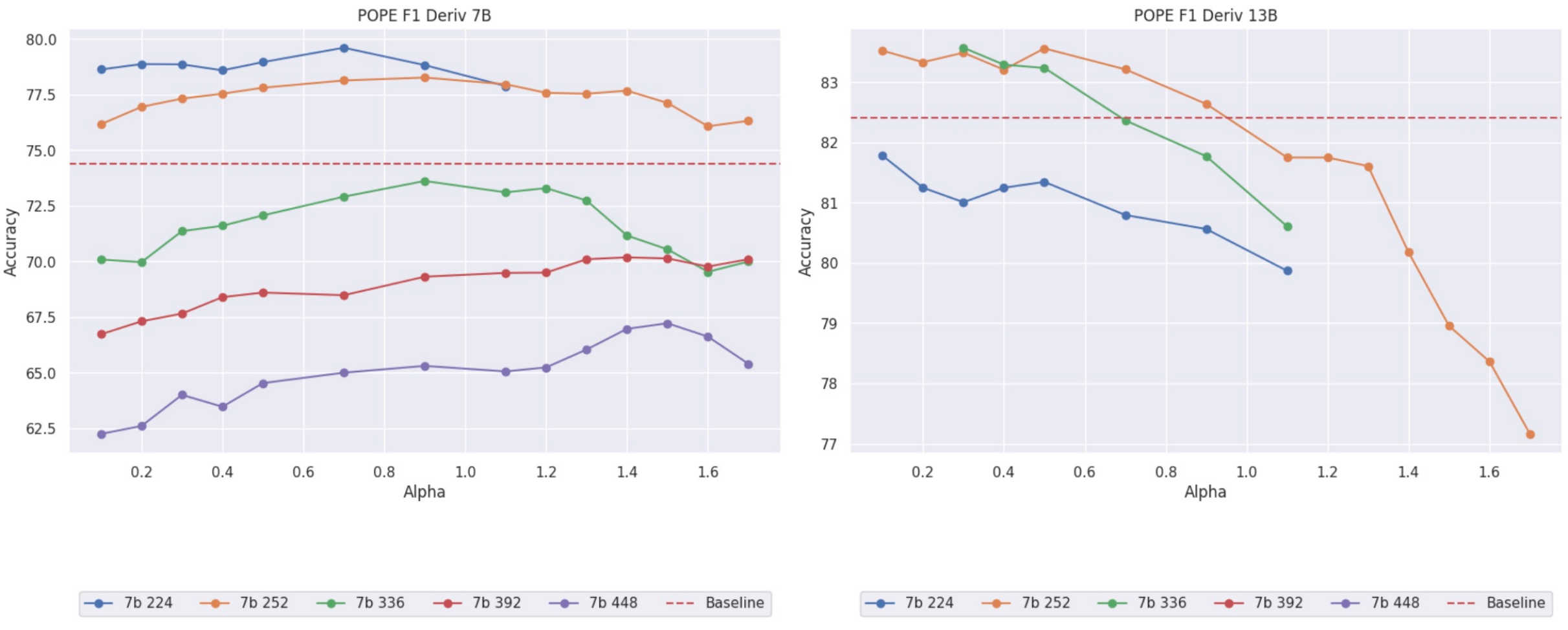}
    \caption{Sweeps on the POPE benchmark \citep{li2023evaluating}. Baselines are indicated by the dashed red line.}
    \label{fig:pope_sweep}
\end{figure}

See Figure~\ref{fig:pope_sweep}. We abandoned sweeps which showed poor performance. POPE evaluations are expensive.

Figure~\ref{fig:pope_sweep}

\subsection{RealWorldQA}
\begin{figure}
    \centering
    \includegraphics[width=0.92\linewidth]{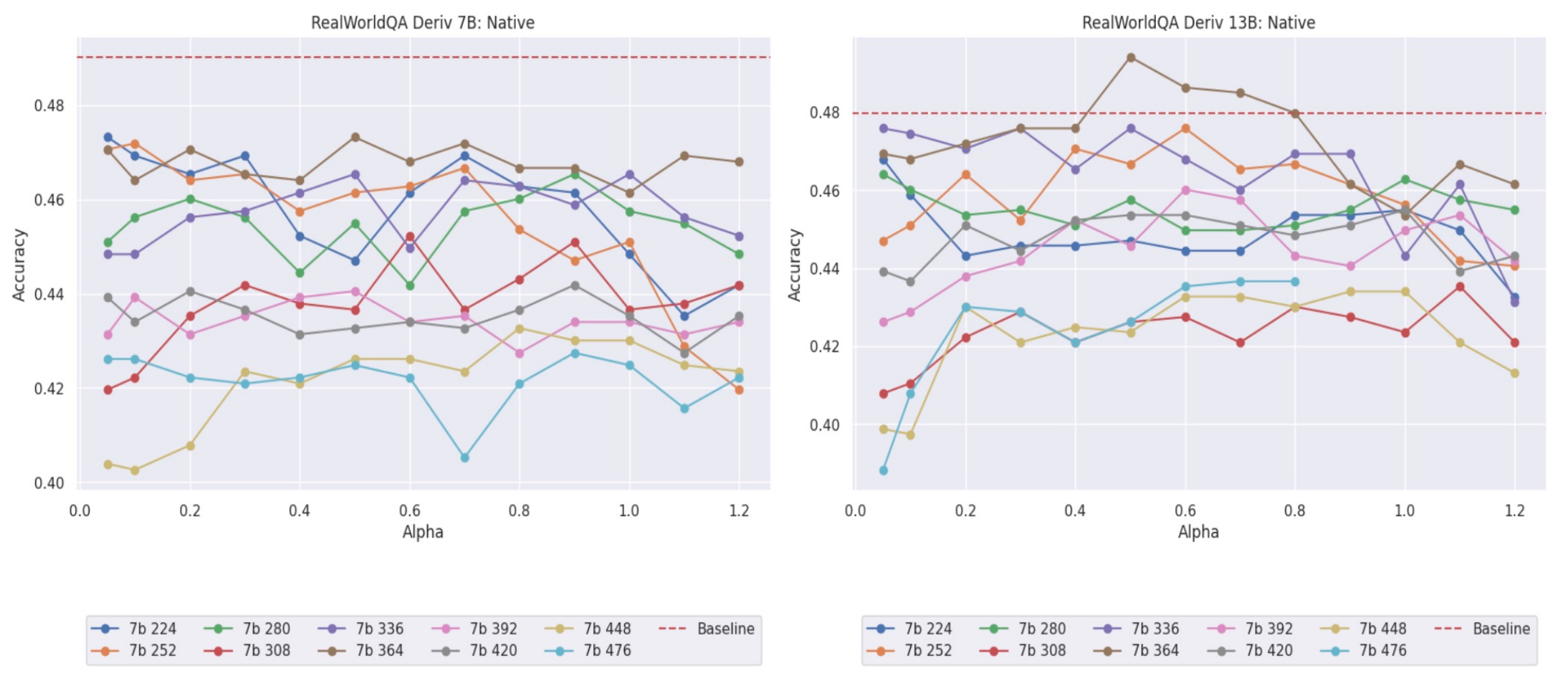}
    \caption{Sweeps on the RealWorldQA benchmark \citep{xai2024realworldqa}. Baselines are indicated by the dashed red line.}
    \label{fig:rqa_sweeps}
\end{figure}

We are able to perform fairly comprehensive sweeps on the RealWorldQA benchmark~\citep{xai2024realworldqa} as the benchmark proves inexpensive to evaluate. The results of our sweeps on the 7B and 13B \modelname model are shown in Figure~\ref{fig:rqa_sweeps}.

\subsection{CV-Bench}
\begin{figure}
    \centering
    \includegraphics[width=0.92\linewidth]{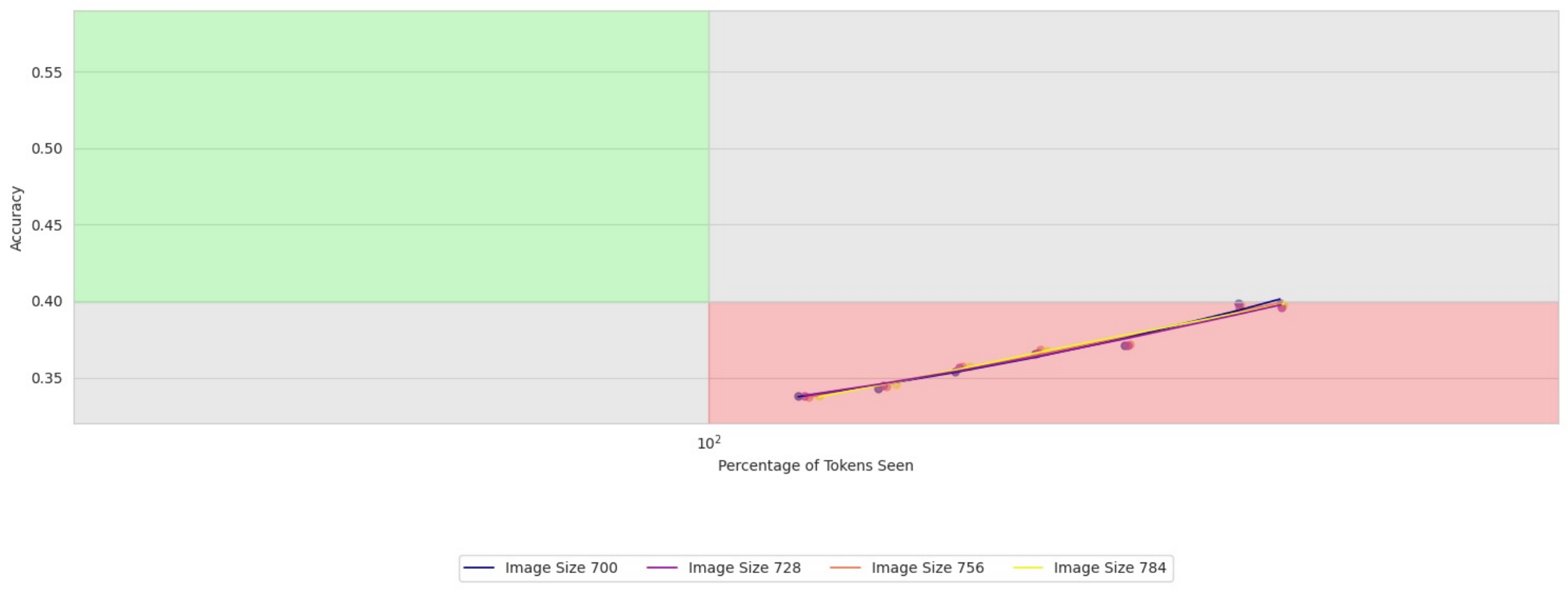}
    \caption{Compute vs. accuracy curves for our sweeps of CV-Bench, 7B, native resolution.}
    \label{fig:cvbench7b}
\end{figure}

CV-Bench \citep{tong2024cambrian} is expensive to run sweeps over. Because of this we searched a relatively small percentage of the search space. We found that performance using the 7B model was best for the larger image sizes (see Figure~\ref{fig:cvbench7b}). We found that the QtP procedure typically led to decreasing performance on CV-Bench, and a preliminary sweep showed us that the 7B model performed best when using the larger image sizes. 

\begin{figure}
    \centering
    \includegraphics[width=0.92\linewidth]{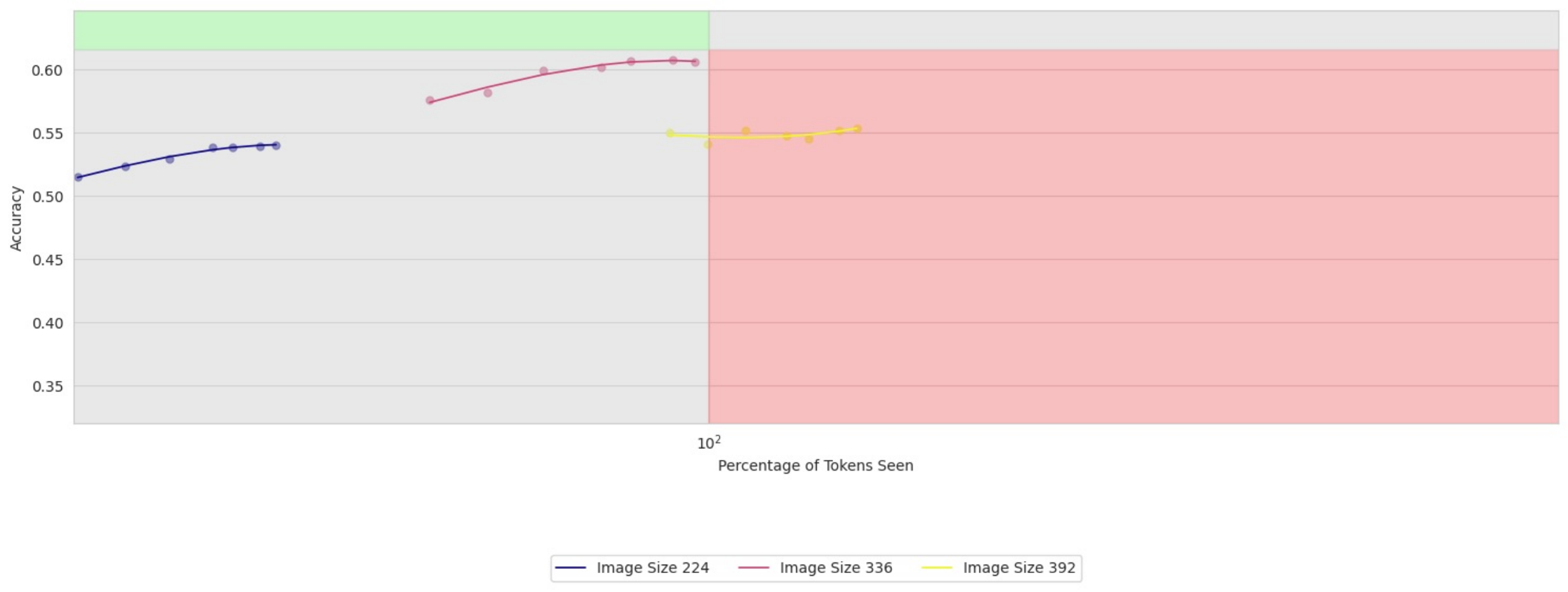}
    \caption{Compute vs. accuracy curves for our sweeps of CV-Bench, 13B, cropped resolution.}
    \label{fig:cvbench13b}
\end{figure}

For 13B the model performed poorly with the native image sizes and we swept crops instead. For this sweep we found smaller images were better, with peak performance occurring when the images matched the pre-training image size, $336\times 336$ (see Figure~\ref{fig:cvbench13b}).

\subsection{$V^*$-Bench}
For the $V^*$ benchmark \citep{wu2024v} we sweep image size and $\alpha$ using our native image resolutions. We did not sweep $V^*$ using cropped images. We sweep image sizes between $224$ and $700$ in steps of $28$. For the derivative selection strategy we sweep $\alpha \in (0.05, 0.1, 0.2, 0.3, 0.4, 0.5, 0.6, 0.7, 0.8, 0.9, 1.0, 1.1, 1.2, 1.3, 1.4, 1.5, 1.7, 1.9, 2.1, 2.5, 3.0)$. For the random selection strategy we sweep $\alpha \in (0.0, 0.05, 0.1, 0.2, 0.3, 0.4, 0.5, 0.6)$. $V^*$ evaluations are the least expensive of our chosen evaluations and therefore we have the most comprehensive sweeps on this benchmark. 

We include the results of our sweeps in color coded tables below. The sweep for the 7B model with the derivative selection strategy can be found in Table~\ref{tab:vstar_deriv_7b}. The sweep for the 7B model with the random selection strategy can be found in Table~\ref{tab:vstar_random_7b}. The sweep for the 13B model with the derivative selection strategy can be found in Table~\ref{tab:vstar_deriv_13b}. The sweep for the 13B model with the random selection strategy can be found in Table~\ref{tab:vstar_random_13b}. 

For the baseline model we sweep 

\begin{table}[]
\caption{Accuracy on $V^*$ for LLaVA-\modelname-7B using derivative selection method. Native resolution.}
\vspace{1em}
\centering
\resizebox{\textwidth}{!}{
\begin{tabular}{ |c|c|c|c|c|c|c|c|c|c|c|c|c|c|c|c|c|c|c|c|c|c| }
\hline
Image Size & $\alpha=0.05$ & $\alpha=0.10$ & $\alpha=0.20$ & $\alpha=0.30$ & $\alpha=0.40$ & $\alpha=0.50$ & $\alpha=0.60$ & $\alpha=0.70$ & $\alpha=0.80$ & $\alpha=0.90$ & $\alpha=1.00$ & $\alpha=1.10$ & $\alpha=1.20$ & $\alpha=1.30$ & $\alpha=1.40$ & $\alpha=1.50$ & $\alpha=1.70$ & $\alpha=1.90$ & $\alpha=2.10$ & $\alpha=2.50$ & $\alpha=3.00$  \\ \hline
224 & \cellcolor[HTML]{c2ffc2} \textbf{45.03}\% & \cellcolor[HTML]{c2ffc2} \textbf{45.03}\% & \cellcolor[HTML]{ceffce} 44.50\% & \cellcolor[HTML]{f2fff2} 42.93\% & \cellcolor[HTML]{e6ffe6} 43.46\% & \cellcolor[HTML]{daffda} 43.98\% & \cellcolor[HTML]{daffda} 43.98\% & \cellcolor[HTML]{f2fff2} 42.93\% &  42.41\% & \cellcolor[HTML]{ffcbcb} 40.84\% & \cellcolor[HTML]{ffcbcb} 40.84\% & \cellcolor[HTML]{ffeeee} 41.88\% &  42.41\% & \cellcolor[HTML]{e6ffe6} 43.46\% & \cellcolor[HTML]{ffeeee} 41.88\% & \cellcolor[HTML]{ffeeee} 41.88\% & \cellcolor[HTML]{c2ffc2} \textbf{45.03}\% &  42.41\% & \cellcolor[HTML]{e6ffe6} 43.46\% & \cellcolor[HTML]{ff5555} 37.17\% & \cellcolor[HTML]{ff9999} 39.27\% \\
252 & \cellcolor[HTML]{9dff9d} 46.60\% & \cellcolor[HTML]{85ff85} 47.64\% & \cellcolor[HTML]{85ff85} 47.64\% & \cellcolor[HTML]{9dff9d} 46.60\% & \cellcolor[HTML]{9dff9d} 46.60\% & \cellcolor[HTML]{85ff85} 47.64\% & \cellcolor[HTML]{79ff79} 48.17\% & \cellcolor[HTML]{48ff48} \textbf{50.26}\% & \cellcolor[HTML]{48ff48} \textbf{50.26}\% & \cellcolor[HTML]{6dff6d} 48.69\% & \cellcolor[HTML]{91ff91} 47.12\% & \cellcolor[HTML]{85ff85} 47.64\% & \cellcolor[HTML]{79ff79} 48.17\% & \cellcolor[HTML]{85ff85} 47.64\% & \cellcolor[HTML]{b6ffb6} 45.55\% & \cellcolor[HTML]{c2ffc2} 45.03\% & \cellcolor[HTML]{c2ffc2} 45.03\% & \cellcolor[HTML]{ffeeee} 41.88\% & \cellcolor[HTML]{f2fff2} 42.93\% & \cellcolor[HTML]{ffbaba} 40.31\% & \cellcolor[HTML]{ff0000} 34.55\% \\
280 & \cellcolor[HTML]{e6ffe6} 43.46\% & \cellcolor[HTML]{f2fff2} 42.93\% & \cellcolor[HTML]{e6ffe6} 43.46\% & \cellcolor[HTML]{f2fff2} 42.93\% & \cellcolor[HTML]{e6ffe6} 43.46\% & \cellcolor[HTML]{f2fff2} 42.93\% &  42.41\% & \cellcolor[HTML]{f2fff2} 42.93\% & \cellcolor[HTML]{f2fff2} 42.93\% & \cellcolor[HTML]{b6ffb6} \textbf{45.55}\% & \cellcolor[HTML]{daffda} 43.98\% & \cellcolor[HTML]{f2fff2} 42.93\% & \cellcolor[HTML]{daffda} 43.98\% & \cellcolor[HTML]{c2ffc2} 45.03\% & \cellcolor[HTML]{ceffce} 44.50\% & \cellcolor[HTML]{daffda} 43.98\% & \cellcolor[HTML]{daffda} 43.98\% & \cellcolor[HTML]{e6ffe6} 43.46\% & \cellcolor[HTML]{e6ffe6} 43.46\% &  42.41\% & \cellcolor[HTML]{ff6666} 37.70\% \\
308 & \cellcolor[HTML]{ceffce} 44.50\% & \cellcolor[HTML]{daffda} 43.98\% & \cellcolor[HTML]{ceffce} 44.50\% & \cellcolor[HTML]{ceffce} 44.50\% & \cellcolor[HTML]{c2ffc2} 45.03\% & \cellcolor[HTML]{c2ffc2} 45.03\% & \cellcolor[HTML]{b6ffb6} 45.55\% & \cellcolor[HTML]{b6ffb6} 45.55\% & \cellcolor[HTML]{b6ffb6} 45.55\% & \cellcolor[HTML]{9dff9d} 46.60\% & \cellcolor[HTML]{9dff9d} 46.60\% & \cellcolor[HTML]{9dff9d} 46.60\% & \cellcolor[HTML]{85ff85} 47.64\% & \cellcolor[HTML]{ceffce} 44.50\% & \cellcolor[HTML]{9dff9d} 46.60\% & \cellcolor[HTML]{79ff79} \textbf{48.17}\% & \cellcolor[HTML]{e6ffe6} 43.46\% & \cellcolor[HTML]{ffdcdc} 41.36\% & \cellcolor[HTML]{ffbaba} 40.31\% & \cellcolor[HTML]{ffdcdc} 41.36\% & \cellcolor[HTML]{ffa9a9} 39.79\% \\
336 & \cellcolor[HTML]{30ff30} \textbf{51.31}\% & \cellcolor[HTML]{30ff30} \textbf{51.31}\% & \cellcolor[HTML]{30ff30} \textbf{51.31}\% & \cellcolor[HTML]{30ff30} \textbf{51.31}\% & \cellcolor[HTML]{54ff54} 49.74\% & \cellcolor[HTML]{91ff91} 47.12\% & \cellcolor[HTML]{91ff91} 47.12\% & \cellcolor[HTML]{91ff91} 47.12\% & \cellcolor[HTML]{91ff91} 47.12\% & \cellcolor[HTML]{79ff79} 48.17\% & \cellcolor[HTML]{61ff61} 49.21\% & \cellcolor[HTML]{6dff6d} 48.69\% & \cellcolor[HTML]{79ff79} 48.17\% & \cellcolor[HTML]{85ff85} 47.64\% & \cellcolor[HTML]{9dff9d} 46.60\% & \cellcolor[HTML]{91ff91} 47.12\% & \cellcolor[HTML]{48ff48} 50.26\% & \cellcolor[HTML]{ceffce} 44.50\% & \cellcolor[HTML]{c2ffc2} 45.03\% & \cellcolor[HTML]{ffcbcb} 40.84\% & \cellcolor[HTML]{ff4444} 36.65\% \\
364 & \cellcolor[HTML]{18ff18} 52.36\% & \cellcolor[HTML]{18ff18} 52.36\% & \cellcolor[HTML]{18ff18} 52.36\% & \cellcolor[HTML]{18ff18} 52.36\% & \cellcolor[HTML]{0cff0c} 52.88\% & \cellcolor[HTML]{00ff00} \underline{\textbf{53.40\%}} & \cellcolor[HTML]{0cff0c} 52.88\% & \cellcolor[HTML]{18ff18} 52.36\% & \cellcolor[HTML]{24ff24} 51.83\% & \cellcolor[HTML]{00ff00} \underline{\textbf{53.40\%}} & \cellcolor[HTML]{30ff30} 51.31\% & \cellcolor[HTML]{3cff3c} 50.79\% & \cellcolor[HTML]{61ff61} 49.21\% & \cellcolor[HTML]{91ff91} 47.12\% & \cellcolor[HTML]{24ff24} 51.83\% & \cellcolor[HTML]{3cff3c} 50.79\% & \cellcolor[HTML]{3cff3c} 50.79\% & \cellcolor[HTML]{85ff85} 47.64\% & \cellcolor[HTML]{9dff9d} 46.60\% & \cellcolor[HTML]{e6ffe6} 43.46\% & \cellcolor[HTML]{ffeeee} 41.88\% \\
392 & \cellcolor[HTML]{6dff6d} \textbf{48.69}\% & \cellcolor[HTML]{85ff85} 47.64\% & \cellcolor[HTML]{6dff6d} \textbf{48.69}\% & \cellcolor[HTML]{85ff85} 47.64\% & \cellcolor[HTML]{85ff85} 47.64\% & \cellcolor[HTML]{91ff91} 47.12\% & \cellcolor[HTML]{85ff85} 47.64\% & \cellcolor[HTML]{b6ffb6} 45.55\% & \cellcolor[HTML]{85ff85} 47.64\% & \cellcolor[HTML]{91ff91} 47.12\% & \cellcolor[HTML]{91ff91} 47.12\% & \cellcolor[HTML]{91ff91} 47.12\% & \cellcolor[HTML]{85ff85} 47.64\% & \cellcolor[HTML]{c2ffc2} 45.03\% & \cellcolor[HTML]{c2ffc2} 45.03\% & \cellcolor[HTML]{9dff9d} 46.60\% & \cellcolor[HTML]{aaffaa} 46.07\% & \cellcolor[HTML]{ceffce} 44.50\% & \cellcolor[HTML]{f2fff2} 42.93\% & \cellcolor[HTML]{daffda} 43.98\% & \cellcolor[HTML]{ff8888} 38.74\% \\
420 & \cellcolor[HTML]{61ff61} 49.21\% & \cellcolor[HTML]{61ff61} 49.21\% & \cellcolor[HTML]{54ff54} 49.74\% & \cellcolor[HTML]{3cff3c} 50.79\% & \cellcolor[HTML]{30ff30} \textbf{51.31}\% & \cellcolor[HTML]{61ff61} 49.21\% & \cellcolor[HTML]{48ff48} 50.26\% & \cellcolor[HTML]{9dff9d} 46.60\% & \cellcolor[HTML]{85ff85} 47.64\% & \cellcolor[HTML]{91ff91} 47.12\% & \cellcolor[HTML]{61ff61} 49.21\% & \cellcolor[HTML]{48ff48} 50.26\% & \cellcolor[HTML]{61ff61} 49.21\% & \cellcolor[HTML]{85ff85} 47.64\% & \cellcolor[HTML]{9dff9d} 46.60\% & \cellcolor[HTML]{b6ffb6} 45.55\% & \cellcolor[HTML]{61ff61} 49.21\% & \cellcolor[HTML]{c2ffc2} 45.03\% &  42.41\% & \cellcolor[HTML]{ffcbcb} 40.84\% & \cellcolor[HTML]{ff6666} 37.70\% \\
448 & \cellcolor[HTML]{30ff30} 51.31\% & \cellcolor[HTML]{3cff3c} 50.79\% & \cellcolor[HTML]{30ff30} 51.31\% & \cellcolor[HTML]{18ff18} 52.36\% & \cellcolor[HTML]{0cff0c} \textbf{52.88}\% & \cellcolor[HTML]{30ff30} 51.31\% & \cellcolor[HTML]{30ff30} 51.31\% & \cellcolor[HTML]{30ff30} 51.31\% & \cellcolor[HTML]{48ff48} 50.26\% & \cellcolor[HTML]{61ff61} 49.21\% & \cellcolor[HTML]{79ff79} 48.17\% & \cellcolor[HTML]{61ff61} 49.21\% & \cellcolor[HTML]{85ff85} 47.64\% & \cellcolor[HTML]{79ff79} 48.17\% & \cellcolor[HTML]{54ff54} 49.74\% & \cellcolor[HTML]{9dff9d} 46.60\% & \cellcolor[HTML]{e6ffe6} 43.46\% & \cellcolor[HTML]{b6ffb6} 45.55\% & \cellcolor[HTML]{9dff9d} 46.60\% & \cellcolor[HTML]{f2fff2} 42.93\% & \cellcolor[HTML]{ffa9a9} 39.79\% \\
476 & \cellcolor[HTML]{c2ffc2} 45.03\% & \cellcolor[HTML]{9dff9d} 46.60\% & \cellcolor[HTML]{9dff9d} 46.60\% & \cellcolor[HTML]{b6ffb6} 45.55\% & \cellcolor[HTML]{b6ffb6} 45.55\% & \cellcolor[HTML]{ceffce} 44.50\% & \cellcolor[HTML]{91ff91} 47.12\% & \cellcolor[HTML]{ceffce} 44.50\% & \cellcolor[HTML]{9dff9d} 46.60\% & \cellcolor[HTML]{9dff9d} 46.60\% & \cellcolor[HTML]{6dff6d} \textbf{48.69}\% & \cellcolor[HTML]{6dff6d} \textbf{48.69}\% & \cellcolor[HTML]{c2ffc2} 45.03\% & \cellcolor[HTML]{b6ffb6} 45.55\% & \cellcolor[HTML]{f2fff2} 42.93\% & \cellcolor[HTML]{ceffce} 44.50\% & \cellcolor[HTML]{b6ffb6} 45.55\% & \cellcolor[HTML]{ffeeee} 41.88\% & \cellcolor[HTML]{ffeeee} 41.88\% & \cellcolor[HTML]{ff7777} 38.22\% & \cellcolor[HTML]{ff4444} 36.65\% \\
504 & \cellcolor[HTML]{48ff48} 50.26\% & \cellcolor[HTML]{48ff48} 50.26\% & \cellcolor[HTML]{30ff30} 51.31\% & \cellcolor[HTML]{0cff0c} \textbf{52.88}\% & \cellcolor[HTML]{24ff24} 51.83\% & \cellcolor[HTML]{30ff30} 51.31\% & \cellcolor[HTML]{18ff18} 52.36\% & \cellcolor[HTML]{18ff18} 52.36\% & \cellcolor[HTML]{24ff24} 51.83\% & \cellcolor[HTML]{54ff54} 49.74\% & \cellcolor[HTML]{30ff30} 51.31\% & \cellcolor[HTML]{6dff6d} 48.69\% & \cellcolor[HTML]{6dff6d} 48.69\% & \cellcolor[HTML]{30ff30} 51.31\% & \cellcolor[HTML]{54ff54} 49.74\% & \cellcolor[HTML]{30ff30} 51.31\% & \cellcolor[HTML]{48ff48} 50.26\% & \cellcolor[HTML]{61ff61} 49.21\% & \cellcolor[HTML]{9dff9d} 46.60\% & \cellcolor[HTML]{daffda} 43.98\% & \cellcolor[HTML]{ffcbcb} 40.84\% \\
532 & \cellcolor[HTML]{6dff6d} 48.69\% & \cellcolor[HTML]{54ff54} 49.74\% & \cellcolor[HTML]{3cff3c} \textbf{50.79}\% & \cellcolor[HTML]{61ff61} 49.21\% & \cellcolor[HTML]{79ff79} 48.17\% & \cellcolor[HTML]{6dff6d} 48.69\% & \cellcolor[HTML]{48ff48} 50.26\% & \cellcolor[HTML]{3cff3c} \textbf{50.79}\% & \cellcolor[HTML]{79ff79} 48.17\% & \cellcolor[HTML]{85ff85} 47.64\% & \cellcolor[HTML]{6dff6d} 48.69\% & \cellcolor[HTML]{54ff54} 49.74\% & \cellcolor[HTML]{54ff54} 49.74\% & \cellcolor[HTML]{79ff79} 48.17\% & \cellcolor[HTML]{b6ffb6} 45.55\% & \cellcolor[HTML]{c2ffc2} 45.03\% & \cellcolor[HTML]{aaffaa} 46.07\% & \cellcolor[HTML]{61ff61} 49.21\% & \cellcolor[HTML]{9dff9d} 46.60\% & \cellcolor[HTML]{f2fff2} 42.93\% & \cellcolor[HTML]{ff7777} 38.22\% \\
560 & \cellcolor[HTML]{aaffaa} 46.07\% & \cellcolor[HTML]{c2ffc2} 45.03\% & \cellcolor[HTML]{9dff9d} 46.60\% & \cellcolor[HTML]{b6ffb6} 45.55\% & \cellcolor[HTML]{ceffce} 44.50\% & \cellcolor[HTML]{f2fff2} 42.93\% & \cellcolor[HTML]{e6ffe6} 43.46\% & \cellcolor[HTML]{c2ffc2} 45.03\% & \cellcolor[HTML]{ceffce} 44.50\% &  42.41\% & \cellcolor[HTML]{f2fff2} 42.93\% & \cellcolor[HTML]{f2fff2} 42.93\% & \cellcolor[HTML]{e6ffe6} 43.46\% & \cellcolor[HTML]{b6ffb6} 45.55\% & \cellcolor[HTML]{9dff9d} 46.60\% & \cellcolor[HTML]{85ff85} \textbf{47.64}\% & \cellcolor[HTML]{aaffaa} 46.07\% & \cellcolor[HTML]{91ff91} 47.12\% & \cellcolor[HTML]{ceffce} 44.50\% & \cellcolor[HTML]{c2ffc2} 45.03\% & \cellcolor[HTML]{ffeeee} 41.88\% \\
588 & \cellcolor[HTML]{91ff91} 47.12\% & \cellcolor[HTML]{9dff9d} 46.60\% & \cellcolor[HTML]{9dff9d} 46.60\% & \cellcolor[HTML]{9dff9d} 46.60\% & \cellcolor[HTML]{79ff79} 48.17\% & \cellcolor[HTML]{79ff79} 48.17\% & \cellcolor[HTML]{61ff61} 49.21\% & \cellcolor[HTML]{6dff6d} 48.69\% & \cellcolor[HTML]{91ff91} 47.12\% & \cellcolor[HTML]{9dff9d} 46.60\% & \cellcolor[HTML]{aaffaa} 46.07\% & \cellcolor[HTML]{91ff91} 47.12\% & \cellcolor[HTML]{6dff6d} 48.69\% & \cellcolor[HTML]{9dff9d} 46.60\% & \cellcolor[HTML]{48ff48} \textbf{50.26}\% & \cellcolor[HTML]{48ff48} \textbf{50.26}\% & \cellcolor[HTML]{aaffaa} 46.07\% & \cellcolor[HTML]{9dff9d} 46.60\% & \cellcolor[HTML]{6dff6d} 48.69\% & \cellcolor[HTML]{f2fff2} 42.93\% & \cellcolor[HTML]{85ff85} 47.64\% \\
616 & \cellcolor[HTML]{ceffce} 44.50\% & \cellcolor[HTML]{f2fff2} 42.93\% & \cellcolor[HTML]{daffda} 43.98\% & \cellcolor[HTML]{b6ffb6} 45.55\% & \cellcolor[HTML]{aaffaa} 46.07\% & \cellcolor[HTML]{ceffce} 44.50\% & \cellcolor[HTML]{aaffaa} 46.07\% & \cellcolor[HTML]{aaffaa} 46.07\% & \cellcolor[HTML]{85ff85} \textbf{47.64}\% & \cellcolor[HTML]{85ff85} \textbf{47.64}\% & \cellcolor[HTML]{85ff85} \textbf{47.64}\% & \cellcolor[HTML]{9dff9d} 46.60\% & \cellcolor[HTML]{aaffaa} 46.07\% & \cellcolor[HTML]{aaffaa} 46.07\% & \cellcolor[HTML]{b6ffb6} 45.55\% & \cellcolor[HTML]{e6ffe6} 43.46\% & \cellcolor[HTML]{ffbaba} 40.31\% & \cellcolor[HTML]{ff6666} 37.70\% & \cellcolor[HTML]{ff5555} 37.17\% & \cellcolor[HTML]{ff6666} 37.70\% & \cellcolor[HTML]{ff6666} 37.70\% \\
644 & \cellcolor[HTML]{daffda} 43.98\% & \cellcolor[HTML]{e6ffe6} 43.46\% & \cellcolor[HTML]{ceffce} 44.50\% & \cellcolor[HTML]{b6ffb6} 45.55\% & \cellcolor[HTML]{aaffaa} 46.07\% & \cellcolor[HTML]{b6ffb6} 45.55\% & \cellcolor[HTML]{aaffaa} 46.07\% & \cellcolor[HTML]{91ff91} 47.12\% & \cellcolor[HTML]{aaffaa} 46.07\% & \cellcolor[HTML]{b6ffb6} 45.55\% & \cellcolor[HTML]{61ff61} \textbf{49.21}\% & \cellcolor[HTML]{85ff85} 47.64\% & \cellcolor[HTML]{c2ffc2} 45.03\% & \cellcolor[HTML]{c2ffc2} 45.03\% & \cellcolor[HTML]{79ff79} 48.17\% & \cellcolor[HTML]{9dff9d} 46.60\% &  42.41\% & \cellcolor[HTML]{ff2222} 35.60\% & \cellcolor[HTML]{ff5555} 37.17\% & \cellcolor[HTML]{ff5555} 37.17\% & \cellcolor[HTML]{ff6666} 37.70\% \\
672 & \cellcolor[HTML]{b6ffb6} 45.55\% & \cellcolor[HTML]{c2ffc2} 45.03\% & \cellcolor[HTML]{b6ffb6} 45.55\% & \cellcolor[HTML]{b6ffb6} 45.55\% & \cellcolor[HTML]{91ff91} 47.12\% & \cellcolor[HTML]{85ff85} 47.64\% & \cellcolor[HTML]{85ff85} 47.64\% & \cellcolor[HTML]{54ff54} \textbf{49.74}\% & \cellcolor[HTML]{91ff91} 47.12\% & \cellcolor[HTML]{c2ffc2} 45.03\% & \cellcolor[HTML]{85ff85} 47.64\% & \cellcolor[HTML]{91ff91} 47.12\% & \cellcolor[HTML]{f2fff2} 42.93\% & \cellcolor[HTML]{aaffaa} 46.07\% & \cellcolor[HTML]{b6ffb6} 45.55\% & \cellcolor[HTML]{f2fff2} 42.93\% & \cellcolor[HTML]{ffbaba} 40.31\% & \cellcolor[HTML]{ff6666} 37.70\% & \cellcolor[HTML]{ff1111} 35.08\% & \cellcolor[HTML]{ff6666} 37.70\% & \cellcolor[HTML]{ff2222} 35.60\% \\
700 & \cellcolor[HTML]{daffda} 43.98\% & \cellcolor[HTML]{c2ffc2} 45.03\% & \cellcolor[HTML]{ceffce} 44.50\% & \cellcolor[HTML]{b6ffb6} 45.55\% & \cellcolor[HTML]{b6ffb6} 45.55\% & \cellcolor[HTML]{91ff91} 47.12\% & \cellcolor[HTML]{9dff9d} 46.60\% & \cellcolor[HTML]{85ff85} \textbf{47.64}\% & \cellcolor[HTML]{91ff91} 47.12\% & \cellcolor[HTML]{c2ffc2} 45.03\% & \cellcolor[HTML]{c2ffc2} 45.03\% & \cellcolor[HTML]{ceffce} 44.50\% & \cellcolor[HTML]{f2fff2} 42.93\% & \cellcolor[HTML]{e6ffe6} 43.46\% & \cellcolor[HTML]{ffbaba} 40.31\% & \cellcolor[HTML]{ceffce} 44.50\% & \cellcolor[HTML]{ffa9a9} 39.79\% & \cellcolor[HTML]{ffeeee} 41.88\% & \cellcolor[HTML]{ff3333} 36.13\% & \cellcolor[HTML]{ff0000} 34.55\% & \cellcolor[HTML]{ff0000} 34.55\% \\
\hline
\end{tabular}
}
\label{tab:vstar_deriv_7b}
\end{table}
\begin{table}[]
\caption{Accuracy on $V^*$ for LLaVA-\modelname-7B using random selection method. Native resolution.}
\vspace{1em}
\centering
\resizebox{\textwidth}{!}{
\begin{tabular}{ |c|c|c|c|c|c|c|c|c| }
\hline
Image Size & $\alpha=0.00$ & $\alpha=0.05$ & $\alpha=0.10$ & $\alpha=0.20$ & $\alpha=0.30$ & $\alpha=0.40$ & $\alpha=0.50$ & $\alpha=0.60$ \\ \hline
224 & \cellcolor[HTML]{a5ffa5} \textbf{46.07}\% & \cellcolor[HTML]{e5ffe5} 43.46\% & \cellcolor[HTML]{ffc2c2} 40.31\% & \cellcolor[HTML]{ffd1d1} 40.84\% & \cellcolor[HTML]{ffd1d1} 40.84\% & \cellcolor[HTML]{ffc2c2} 40.31\% & \cellcolor[HTML]{ff0f0f} 34.03\% & \cellcolor[HTML]{ff3b3b} 35.60\% \\
252 & \cellcolor[HTML]{bfffbf} 45.03\% & \cellcolor[HTML]{8cff8c} \textbf{47.12}\% & \cellcolor[HTML]{b2ffb2} 45.55\% & \cellcolor[HTML]{ffe0e0} 41.36\% & \cellcolor[HTML]{ffd1d1} 40.84\% & \cellcolor[HTML]{ff3b3b} 35.60\% & \cellcolor[HTML]{ff5959} 36.65\% & \cellcolor[HTML]{ff2c2c} 35.08\% \\
280 & \cellcolor[HTML]{d8ffd8} 43.98\% & \cellcolor[HTML]{cbffcb} 44.50\% & \cellcolor[HTML]{e5ffe5} 43.46\% & \cellcolor[HTML]{bfffbf} \textbf{45.03}\% & \cellcolor[HTML]{ffb3b3} 39.79\% & \cellcolor[HTML]{ffc2c2} 40.31\% & \cellcolor[HTML]{f2fff2} 42.93\% & \cellcolor[HTML]{ffd1d1} 40.84\% \\
308 & \cellcolor[HTML]{b2ffb2} \textbf{45.55}\% &  42.41\% & \cellcolor[HTML]{ffe0e0} 41.36\% & \cellcolor[HTML]{ffd1d1} 40.84\% & \cellcolor[HTML]{ff7878} 37.70\% & \cellcolor[HTML]{ff7878} 37.70\% & \cellcolor[HTML]{ff9696} 38.74\% & \cellcolor[HTML]{ff6868} 37.17\% \\
336 & \cellcolor[HTML]{26ff26} \textbf{51.31}\% & \cellcolor[HTML]{98ff98} 46.60\% & \cellcolor[HTML]{b2ffb2} 45.55\% & \cellcolor[HTML]{d8ffd8} 43.98\% & \cellcolor[HTML]{ffe0e0} 41.36\% & \cellcolor[HTML]{ff3b3b} 35.60\% & \cellcolor[HTML]{ff9696} 38.74\% & \cellcolor[HTML]{ff6868} 37.17\% \\
364 & \cellcolor[HTML]{00ff00} \underline{\textbf{52.88\%}} & \cellcolor[HTML]{7fff7f} 47.64\% & \cellcolor[HTML]{19ff19} 51.83\% & \cellcolor[HTML]{cbffcb} 44.50\% & \cellcolor[HTML]{bfffbf} 45.03\% & \cellcolor[HTML]{ff4a4a} 36.13\% & \cellcolor[HTML]{cbffcb} 44.50\% & \cellcolor[HTML]{ff9696} 38.74\% \\
392 & \cellcolor[HTML]{65ff65} 48.69\% & \cellcolor[HTML]{59ff59} \textbf{49.21}\% & \cellcolor[HTML]{98ff98} 46.60\% & \cellcolor[HTML]{f2fff2} 42.93\% &  42.41\% & \cellcolor[HTML]{ffefef} 41.88\% & \cellcolor[HTML]{e5ffe5} 43.46\% & \cellcolor[HTML]{ff0000} 33.51\% \\
420 & \cellcolor[HTML]{4cff4c} \textbf{49.74}\% & \cellcolor[HTML]{8cff8c} 47.12\% & \cellcolor[HTML]{7fff7f} 47.64\% & \cellcolor[HTML]{8cff8c} 47.12\% & \cellcolor[HTML]{72ff72} 48.17\% & \cellcolor[HTML]{ff6868} 37.17\% & \cellcolor[HTML]{ff5959} 36.65\% & \cellcolor[HTML]{ff7878} 37.70\% \\
448 & \cellcolor[HTML]{00ff00} \underline{\textbf{52.88\%}} & \cellcolor[HTML]{32ff32} 50.79\% & \cellcolor[HTML]{b2ffb2} 45.55\% & \cellcolor[HTML]{e5ffe5} 43.46\% & \cellcolor[HTML]{ffd1d1} 40.84\% & \cellcolor[HTML]{ff6868} 37.17\% & \cellcolor[HTML]{ffa5a5} 39.27\% & \cellcolor[HTML]{ff8787} 38.22\% \\
476 & \cellcolor[HTML]{a5ffa5} \textbf{46.07}\% & \cellcolor[HTML]{bfffbf} 45.03\% & \cellcolor[HTML]{f2fff2} 42.93\% & \cellcolor[HTML]{d8ffd8} 43.98\% & \cellcolor[HTML]{f2fff2} 42.93\% & \cellcolor[HTML]{ffd1d1} 40.84\% & \cellcolor[HTML]{ff0000} 33.51\% & \cellcolor[HTML]{ff2c2c} 35.08\% \\
504 & \cellcolor[HTML]{4cff4c} \textbf{49.74}\% & \cellcolor[HTML]{7fff7f} 47.64\% & \cellcolor[HTML]{72ff72} 48.17\% & \cellcolor[HTML]{4cff4c} \textbf{49.74}\% &  42.41\% & \cellcolor[HTML]{cbffcb} 44.50\% & \cellcolor[HTML]{ff7878} 37.70\% & \cellcolor[HTML]{ffc2c2} 40.31\% \\
532 & \cellcolor[HTML]{72ff72} 48.17\% & \cellcolor[HTML]{59ff59} \textbf{49.21}\% & \cellcolor[HTML]{8cff8c} 47.12\% & \cellcolor[HTML]{72ff72} 48.17\% & \cellcolor[HTML]{f2fff2} 42.93\% &  42.41\% & \cellcolor[HTML]{ffa5a5} 39.27\% & \cellcolor[HTML]{ffefef} 41.88\% \\
560 & \cellcolor[HTML]{a5ffa5} 46.07\% & \cellcolor[HTML]{b2ffb2} 45.55\% & \cellcolor[HTML]{d8ffd8} 43.98\% & \cellcolor[HTML]{65ff65} \textbf{48.69}\% & \cellcolor[HTML]{ffefef} 41.88\% &  42.41\% & \cellcolor[HTML]{e5ffe5} 43.46\% & \cellcolor[HTML]{f2fff2} 42.93\% \\
588 & \cellcolor[HTML]{7fff7f} 47.64\% & \cellcolor[HTML]{98ff98} 46.60\% & \cellcolor[HTML]{4cff4c} \textbf{49.74}\% & \cellcolor[HTML]{b2ffb2} 45.55\% & \cellcolor[HTML]{59ff59} 49.21\% & \cellcolor[HTML]{a5ffa5} 46.07\% & \cellcolor[HTML]{d8ffd8} 43.98\% & \cellcolor[HTML]{a5ffa5} 46.07\% \\
616 & \cellcolor[HTML]{cbffcb} 44.50\% & \cellcolor[HTML]{cbffcb} 44.50\% & \cellcolor[HTML]{ffefef} 41.88\% & \cellcolor[HTML]{b2ffb2} \textbf{45.55}\% & \cellcolor[HTML]{ffb3b3} 39.79\% & \cellcolor[HTML]{ff9696} 38.74\% & \cellcolor[HTML]{ffc2c2} 40.31\% & \cellcolor[HTML]{ff7878} 37.70\% \\
644 & \cellcolor[HTML]{f2fff2} 42.93\% & \cellcolor[HTML]{a5ffa5} 46.07\% & \cellcolor[HTML]{b2ffb2} 45.55\% & \cellcolor[HTML]{72ff72} \textbf{48.17}\% & \cellcolor[HTML]{ff3b3b} 35.60\% &  42.41\% & \cellcolor[HTML]{ffb3b3} 39.79\% & \cellcolor[HTML]{ff7878} 37.70\% \\
672 & \cellcolor[HTML]{bfffbf} 45.03\% & \cellcolor[HTML]{bfffbf} 45.03\% & \cellcolor[HTML]{7fff7f} \textbf{47.64}\% & \cellcolor[HTML]{ffefef} 41.88\% & \cellcolor[HTML]{ffa5a5} 39.27\% & \cellcolor[HTML]{ffa5a5} 39.27\% & \cellcolor[HTML]{ffa5a5} 39.27\% & \cellcolor[HTML]{ff0f0f} 34.03\% \\
700 & \cellcolor[HTML]{e5ffe5} 43.46\% & \cellcolor[HTML]{98ff98} \textbf{46.60}\% & \cellcolor[HTML]{a5ffa5} 46.07\% & \cellcolor[HTML]{ffb3b3} 39.79\% & \cellcolor[HTML]{ffefef} 41.88\% & \cellcolor[HTML]{ffc2c2} 40.31\% & \cellcolor[HTML]{ffb3b3} 39.79\% & \cellcolor[HTML]{ff2c2c} 35.08\% \\
\hline
\end{tabular}
}
\label{tab:vstar_random_7b}
\end{table}
\begin{table}[]
\caption{Accuracy on $V^*$ for LLaVA-\modelname-13B using derivative selection method. Native resolution.}
\vspace{1em}
\centering
\resizebox{\textwidth}{!}{
\begin{tabular}{ |c|c|c|c|c|c|c|c|c|c|c|c|c|c|c|c|c|c|c|c|c|c| }
\hline
Image Size & $\alpha=0.05$ & $\alpha=0.10$ & $\alpha=0.20$ & $\alpha=0.30$ & $\alpha=0.40$ & $\alpha=0.50$ & $\alpha=0.60$ & $\alpha=0.70$ & $\alpha=0.80$ & $\alpha=0.90$ & $\alpha=1.00$ & $\alpha=1.10$ & $\alpha=1.20$ & $\alpha=1.30$ & $\alpha=1.40$ & $\alpha=1.50$ & $\alpha=1.70$ & $\alpha=1.90$ & $\alpha=2.10$ & $\alpha=2.50$ & $\alpha=3.00$ \\ \hline
224 & \cellcolor[HTML]{9cff9c} 50.26\% & \cellcolor[HTML]{f5fff5} 45.55\% & \cellcolor[HTML]{baffba} 48.69\% & \cellcolor[HTML]{ebffeb} 46.07\% & \cellcolor[HTML]{89ff89} \textbf{51.31}\% & \cellcolor[HTML]{b0ffb0} 49.21\% & \cellcolor[HTML]{cdffcd} 47.64\% & \cellcolor[HTML]{e1ffe1} 46.60\% & \cellcolor[HTML]{baffba} 48.69\% &  45.03\% & \cellcolor[HTML]{cdffcd} 47.64\% & \cellcolor[HTML]{baffba} 48.69\% & \cellcolor[HTML]{c4ffc4} 48.17\% & \cellcolor[HTML]{ebffeb} 46.07\% & \cellcolor[HTML]{fff1f1} 44.50\% & \cellcolor[HTML]{d7ffd7} 47.12\% & \cellcolor[HTML]{ffaeae} 41.88\% & \cellcolor[HTML]{cdffcd} 47.64\% & \cellcolor[HTML]{ffc9c9} 42.93\% & \cellcolor[HTML]{ffa1a1} 41.36\% & \cellcolor[HTML]{ff0d0d} 35.60\% \\
252 & \cellcolor[HTML]{a6ffa6} 49.74\% & \cellcolor[HTML]{9cff9c} 50.26\% & \cellcolor[HTML]{89ff89} 51.31\% & \cellcolor[HTML]{75ff75} 52.36\% & \cellcolor[HTML]{93ff93} 50.79\% & \cellcolor[HTML]{6bff6b} 52.88\% & \cellcolor[HTML]{44ff44} \textbf{54.97}\% & \cellcolor[HTML]{6bff6b} 52.88\% & \cellcolor[HTML]{7fff7f} 51.83\% & \cellcolor[HTML]{b0ffb0} 49.21\% & \cellcolor[HTML]{58ff58} 53.93\% & \cellcolor[HTML]{75ff75} 52.36\% & \cellcolor[HTML]{7fff7f} 51.83\% & \cellcolor[HTML]{9cff9c} 50.26\% & \cellcolor[HTML]{f5fff5} 45.55\% & \cellcolor[HTML]{d7ffd7} 47.12\% & \cellcolor[HTML]{ffe4e4} 43.98\% & \cellcolor[HTML]{ffc9c9} 42.93\% & \cellcolor[HTML]{ff8686} 40.31\% & \cellcolor[HTML]{ff6b6b} 39.27\% & \cellcolor[HTML]{ff2828} 36.65\% \\
280 & \cellcolor[HTML]{baffba} 48.69\% & \cellcolor[HTML]{a6ffa6} 49.74\% & \cellcolor[HTML]{baffba} 48.69\% & \cellcolor[HTML]{ebffeb} 46.07\% & \cellcolor[HTML]{f5fff5} 45.55\% & \cellcolor[HTML]{9cff9c} 50.26\% & \cellcolor[HTML]{b0ffb0} 49.21\% & \cellcolor[HTML]{cdffcd} 47.64\% & \cellcolor[HTML]{b0ffb0} 49.21\% & \cellcolor[HTML]{c4ffc4} 48.17\% & \cellcolor[HTML]{d7ffd7} 47.12\% & \cellcolor[HTML]{93ff93} \textbf{50.79}\% & \cellcolor[HTML]{9cff9c} 50.26\% & \cellcolor[HTML]{a6ffa6} 49.74\% & \cellcolor[HTML]{cdffcd} 47.64\% & \cellcolor[HTML]{d7ffd7} 47.12\% & \cellcolor[HTML]{fff1f1} 44.50\% & \cellcolor[HTML]{ffe4e4} 43.98\% & \cellcolor[HTML]{ffa1a1} 41.36\% & \cellcolor[HTML]{ffbbbb} 42.41\% & \cellcolor[HTML]{ff8686} 40.31\% \\
308 & \cellcolor[HTML]{d7ffd7} 47.12\% & \cellcolor[HTML]{baffba} 48.69\% & \cellcolor[HTML]{d7ffd7} 47.12\% & \cellcolor[HTML]{c4ffc4} 48.17\% & \cellcolor[HTML]{d7ffd7} 47.12\% & \cellcolor[HTML]{b0ffb0} \textbf{49.21}\% & \cellcolor[HTML]{b0ffb0} \textbf{49.21}\% & \cellcolor[HTML]{cdffcd} 47.64\% &  45.03\% & \cellcolor[HTML]{baffba} 48.69\% & \cellcolor[HTML]{ebffeb} 46.07\% & \cellcolor[HTML]{cdffcd} 47.64\% & \cellcolor[HTML]{cdffcd} 47.64\% & \cellcolor[HTML]{f5fff5} 45.55\% & \cellcolor[HTML]{d7ffd7} 47.12\% &  45.03\% & \cellcolor[HTML]{ffc9c9} 42.93\% & \cellcolor[HTML]{ffa1a1} 41.36\% & \cellcolor[HTML]{ff7878} 39.79\% & \cellcolor[HTML]{ffa1a1} 41.36\% & \cellcolor[HTML]{ff2828} 36.65\% \\
336 & \cellcolor[HTML]{93ff93} 50.79\% & \cellcolor[HTML]{93ff93} 50.79\% & \cellcolor[HTML]{7fff7f} 51.83\% & \cellcolor[HTML]{62ff62} 53.40\% & \cellcolor[HTML]{6bff6b} 52.88\% & \cellcolor[HTML]{75ff75} 52.36\% & \cellcolor[HTML]{13ff13} \textbf{57.59}\% & \cellcolor[HTML]{4eff4e} 54.45\% & \cellcolor[HTML]{75ff75} 52.36\% & \cellcolor[HTML]{3aff3a} 55.50\% & \cellcolor[HTML]{58ff58} 53.93\% & \cellcolor[HTML]{44ff44} 54.97\% & \cellcolor[HTML]{13ff13} \textbf{57.59}\% & \cellcolor[HTML]{3aff3a} 55.50\% & \cellcolor[HTML]{93ff93} 50.79\% & \cellcolor[HTML]{a6ffa6} 49.74\% & \cellcolor[HTML]{a6ffa6} 49.74\% & \cellcolor[HTML]{baffba} 48.69\% & \cellcolor[HTML]{ffd6d6} 43.46\% & \cellcolor[HTML]{fff1f1} 44.50\% & \cellcolor[HTML]{ff4343} 37.70\% \\
364 & \cellcolor[HTML]{62ff62} 53.40\% & \cellcolor[HTML]{3aff3a} 55.50\% & \cellcolor[HTML]{27ff27} \textbf{56.54}\% & \cellcolor[HTML]{3aff3a} 55.50\% & \cellcolor[HTML]{4eff4e} 54.45\% & \cellcolor[HTML]{4eff4e} 54.45\% & \cellcolor[HTML]{44ff44} 54.97\% & \cellcolor[HTML]{58ff58} 53.93\% & \cellcolor[HTML]{58ff58} 53.93\% & \cellcolor[HTML]{58ff58} 53.93\% & \cellcolor[HTML]{58ff58} 53.93\% & \cellcolor[HTML]{9cff9c} 50.26\% & \cellcolor[HTML]{c4ffc4} 48.17\% & \cellcolor[HTML]{9cff9c} 50.26\% & \cellcolor[HTML]{9cff9c} 50.26\% & \cellcolor[HTML]{9cff9c} 50.26\% & \cellcolor[HTML]{89ff89} 51.31\% & \cellcolor[HTML]{a6ffa6} 49.74\% & \cellcolor[HTML]{cdffcd} 47.64\% & \cellcolor[HTML]{ffaeae} 41.88\% & \cellcolor[HTML]{ff5d5d} 38.74\% \\
392 & \cellcolor[HTML]{89ff89} 51.31\% & \cellcolor[HTML]{3aff3a} \textbf{55.50}\% & \cellcolor[HTML]{62ff62} 53.40\% & \cellcolor[HTML]{4eff4e} 54.45\% & \cellcolor[HTML]{4eff4e} 54.45\% & \cellcolor[HTML]{6bff6b} 52.88\% & \cellcolor[HTML]{89ff89} 51.31\% & \cellcolor[HTML]{4eff4e} 54.45\% & \cellcolor[HTML]{75ff75} 52.36\% & \cellcolor[HTML]{4eff4e} 54.45\% & \cellcolor[HTML]{93ff93} 50.79\% & \cellcolor[HTML]{9cff9c} 50.26\% & \cellcolor[HTML]{75ff75} 52.36\% & \cellcolor[HTML]{7fff7f} 51.83\% & \cellcolor[HTML]{89ff89} 51.31\% & \cellcolor[HTML]{7fff7f} 51.83\% & \cellcolor[HTML]{7fff7f} 51.83\% & \cellcolor[HTML]{ebffeb} 46.07\% & \cellcolor[HTML]{fff1f1} 44.50\% & \cellcolor[HTML]{ffc9c9} 42.93\% & \cellcolor[HTML]{ff1a1a} 36.13\% \\
420 & \cellcolor[HTML]{a6ffa6} 49.74\% & \cellcolor[HTML]{6bff6b} 52.88\% & \cellcolor[HTML]{62ff62} 53.40\% & \cellcolor[HTML]{6bff6b} 52.88\% & \cellcolor[HTML]{3aff3a} \textbf{55.50}\% & \cellcolor[HTML]{6bff6b} 52.88\% & \cellcolor[HTML]{62ff62} 53.40\% & \cellcolor[HTML]{6bff6b} 52.88\% & \cellcolor[HTML]{89ff89} 51.31\% & \cellcolor[HTML]{6bff6b} 52.88\% & \cellcolor[HTML]{7fff7f} 51.83\% & \cellcolor[HTML]{89ff89} 51.31\% & \cellcolor[HTML]{6bff6b} 52.88\% & \cellcolor[HTML]{6bff6b} 52.88\% & \cellcolor[HTML]{7fff7f} 51.83\% & \cellcolor[HTML]{a6ffa6} 49.74\% & \cellcolor[HTML]{c4ffc4} 48.17\% & \cellcolor[HTML]{ebffeb} 46.07\% & \cellcolor[HTML]{fff1f1} 44.50\% & \cellcolor[HTML]{ff7878} 39.79\% & \cellcolor[HTML]{ff4343} 37.70\% \\
448 & \cellcolor[HTML]{93ff93} 50.79\% & \cellcolor[HTML]{a6ffa6} 49.74\% & \cellcolor[HTML]{93ff93} 50.79\% & \cellcolor[HTML]{89ff89} \textbf{51.31}\% & \cellcolor[HTML]{cdffcd} 47.64\% & \cellcolor[HTML]{d7ffd7} 47.12\% & \cellcolor[HTML]{baffba} 48.69\% & \cellcolor[HTML]{89ff89} \textbf{51.31}\% & \cellcolor[HTML]{89ff89} \textbf{51.31}\% & \cellcolor[HTML]{b0ffb0} 49.21\% & \cellcolor[HTML]{89ff89} \textbf{51.31}\% & \cellcolor[HTML]{89ff89} \textbf{51.31}\% & \cellcolor[HTML]{9cff9c} 50.26\% & \cellcolor[HTML]{cdffcd} 47.64\% & \cellcolor[HTML]{c4ffc4} 48.17\% & \cellcolor[HTML]{93ff93} 50.79\% & \cellcolor[HTML]{baffba} 48.69\% & \cellcolor[HTML]{c4ffc4} 48.17\% & \cellcolor[HTML]{f5fff5} 45.55\% & \cellcolor[HTML]{ffbbbb} 42.41\% & \cellcolor[HTML]{ff4343} 37.70\% \\
476 & \cellcolor[HTML]{b0ffb0} 49.21\% & \cellcolor[HTML]{89ff89} 51.31\% & \cellcolor[HTML]{89ff89} 51.31\% & \cellcolor[HTML]{a6ffa6} 49.74\% & \cellcolor[HTML]{b0ffb0} 49.21\% & \cellcolor[HTML]{a6ffa6} 49.74\% & \cellcolor[HTML]{7fff7f} 51.83\% & \cellcolor[HTML]{75ff75} 52.36\% & \cellcolor[HTML]{75ff75} 52.36\% & \cellcolor[HTML]{3aff3a} 55.50\% & \cellcolor[HTML]{1dff1d} 57.07\% & \cellcolor[HTML]{13ff13} 57.59\% & \cellcolor[HTML]{1dff1d} 57.07\% & \cellcolor[HTML]{00ff00} \underline{\textbf{58.64\%}} & \cellcolor[HTML]{13ff13} 57.59\% & \cellcolor[HTML]{27ff27} 56.54\% & \cellcolor[HTML]{a6ffa6} 49.74\% & \cellcolor[HTML]{f5fff5} 45.55\% & \cellcolor[HTML]{d7ffd7} 47.12\% & \cellcolor[HTML]{ff9393} 40.84\% & \cellcolor[HTML]{ff6b6b} 39.27\% \\
504 & \cellcolor[HTML]{c4ffc4} 48.17\% & \cellcolor[HTML]{c4ffc4} 48.17\% & \cellcolor[HTML]{a6ffa6} 49.74\% & \cellcolor[HTML]{89ff89} 51.31\% & \cellcolor[HTML]{a6ffa6} 49.74\% & \cellcolor[HTML]{93ff93} 50.79\% & \cellcolor[HTML]{7fff7f} 51.83\% & \cellcolor[HTML]{a6ffa6} 49.74\% & \cellcolor[HTML]{93ff93} 50.79\% & \cellcolor[HTML]{a6ffa6} 49.74\% & \cellcolor[HTML]{7fff7f} 51.83\% & \cellcolor[HTML]{62ff62} 53.40\% & \cellcolor[HTML]{6bff6b} 52.88\% & \cellcolor[HTML]{6bff6b} 52.88\% & \cellcolor[HTML]{4eff4e} \textbf{54.45}\% & \cellcolor[HTML]{75ff75} 52.36\% & \cellcolor[HTML]{75ff75} 52.36\% & \cellcolor[HTML]{9cff9c} 50.26\% & \cellcolor[HTML]{baffba} 48.69\% & \cellcolor[HTML]{ffaeae} 41.88\% & \cellcolor[HTML]{ff8686} 40.31\% \\
532 & \cellcolor[HTML]{d7ffd7} 47.12\% & \cellcolor[HTML]{f5fff5} 45.55\% & \cellcolor[HTML]{baffba} 48.69\% & \cellcolor[HTML]{baffba} 48.69\% & \cellcolor[HTML]{ebffeb} 46.07\% & \cellcolor[HTML]{e1ffe1} 46.60\% & \cellcolor[HTML]{e1ffe1} 46.60\% & \cellcolor[HTML]{baffba} 48.69\% & \cellcolor[HTML]{c4ffc4} 48.17\% & \cellcolor[HTML]{b0ffb0} 49.21\% & \cellcolor[HTML]{baffba} 48.69\% & \cellcolor[HTML]{baffba} 48.69\% & \cellcolor[HTML]{baffba} 48.69\% & \cellcolor[HTML]{b0ffb0} 49.21\% & \cellcolor[HTML]{b0ffb0} 49.21\% & \cellcolor[HTML]{9cff9c} 50.26\% & \cellcolor[HTML]{a6ffa6} 49.74\% & \cellcolor[HTML]{89ff89} \textbf{51.31}\% & \cellcolor[HTML]{cdffcd} 47.64\% & \cellcolor[HTML]{fff1f1} 44.50\% & \cellcolor[HTML]{ffaeae} 41.88\% \\
560 & \cellcolor[HTML]{e1ffe1} 46.60\% & \cellcolor[HTML]{cdffcd} 47.64\% & \cellcolor[HTML]{baffba} 48.69\% & \cellcolor[HTML]{a6ffa6} 49.74\% & \cellcolor[HTML]{9cff9c} 50.26\% & \cellcolor[HTML]{9cff9c} 50.26\% & \cellcolor[HTML]{a6ffa6} 49.74\% & \cellcolor[HTML]{b0ffb0} 49.21\% & \cellcolor[HTML]{75ff75} \textbf{52.36}\% & \cellcolor[HTML]{7fff7f} 51.83\% & \cellcolor[HTML]{7fff7f} 51.83\% & \cellcolor[HTML]{baffba} 48.69\% & \cellcolor[HTML]{c4ffc4} 48.17\% & \cellcolor[HTML]{b0ffb0} 49.21\% & \cellcolor[HTML]{ebffeb} 46.07\% & \cellcolor[HTML]{cdffcd} 47.64\% & \cellcolor[HTML]{c4ffc4} 48.17\% & \cellcolor[HTML]{89ff89} 51.31\% & \cellcolor[HTML]{baffba} 48.69\% & \cellcolor[HTML]{cdffcd} 47.64\% & \cellcolor[HTML]{ffaeae} 41.88\% \\
588 & \cellcolor[HTML]{ffe4e4} 43.98\% & \cellcolor[HTML]{ffe4e4} 43.98\% & \cellcolor[HTML]{fff1f1} 44.50\% & \cellcolor[HTML]{d7ffd7} 47.12\% & \cellcolor[HTML]{e1ffe1} 46.60\% & \cellcolor[HTML]{baffba} 48.69\% & \cellcolor[HTML]{c4ffc4} 48.17\% & \cellcolor[HTML]{ebffeb} 46.07\% & \cellcolor[HTML]{ebffeb} 46.07\% & \cellcolor[HTML]{a6ffa6} \textbf{49.74}\% & \cellcolor[HTML]{f5fff5} 45.55\% &  45.03\% & \cellcolor[HTML]{e1ffe1} 46.60\% & \cellcolor[HTML]{ebffeb} 46.07\% & \cellcolor[HTML]{e1ffe1} 46.60\% & \cellcolor[HTML]{cdffcd} 47.64\% & \cellcolor[HTML]{ffe4e4} 43.98\% & \cellcolor[HTML]{d7ffd7} 47.12\% & \cellcolor[HTML]{ffc9c9} 42.93\% & \cellcolor[HTML]{ffa1a1} 41.36\% & \cellcolor[HTML]{ffaeae} 41.88\% \\
616 & \cellcolor[HTML]{ff9393} 40.84\% & \cellcolor[HTML]{ffbbbb} 42.41\% & \cellcolor[HTML]{ffd6d6} 43.46\% & \cellcolor[HTML]{ff9393} 40.84\% & \cellcolor[HTML]{ebffeb} 46.07\% &  45.03\% & \cellcolor[HTML]{f5fff5} 45.55\% & \cellcolor[HTML]{d7ffd7} 47.12\% & \cellcolor[HTML]{e1ffe1} 46.60\% & \cellcolor[HTML]{a6ffa6} \textbf{49.74}\% & \cellcolor[HTML]{cdffcd} 47.64\% & \cellcolor[HTML]{e1ffe1} 46.60\% & \cellcolor[HTML]{d7ffd7} 47.12\% & \cellcolor[HTML]{cdffcd} 47.64\% & \cellcolor[HTML]{ffbbbb} 42.41\% & \cellcolor[HTML]{ffe4e4} 43.98\% & \cellcolor[HTML]{ff4343} 37.70\% & \cellcolor[HTML]{ff8686} 40.31\% & \cellcolor[HTML]{ff3535} 37.17\% & \cellcolor[HTML]{ff4343} 37.70\% & \cellcolor[HTML]{ff5050} 38.22\% \\
644 & \cellcolor[HTML]{ffbbbb} 42.41\% & \cellcolor[HTML]{ffbbbb} 42.41\% & \cellcolor[HTML]{ffe4e4} 43.98\% & \cellcolor[HTML]{ffe4e4} 43.98\% & \cellcolor[HTML]{ffc9c9} 42.93\% & \cellcolor[HTML]{f5fff5} 45.55\% & \cellcolor[HTML]{e1ffe1} \textbf{46.60}\% & \cellcolor[HTML]{fff1f1} 44.50\% & \cellcolor[HTML]{ebffeb} 46.07\% & \cellcolor[HTML]{e1ffe1} \textbf{46.60}\% &  45.03\% & \cellcolor[HTML]{f5fff5} 45.55\% & \cellcolor[HTML]{ffe4e4} 43.98\% & \cellcolor[HTML]{f5fff5} 45.55\% & \cellcolor[HTML]{fff1f1} 44.50\% & \cellcolor[HTML]{ffd6d6} 43.46\% & \cellcolor[HTML]{fff1f1} 44.50\% & \cellcolor[HTML]{ffbbbb} 42.41\% & \cellcolor[HTML]{ff6b6b} 39.27\% & \cellcolor[HTML]{ff2828} 36.65\% & \cellcolor[HTML]{ff3535} 37.17\% \\
672 & \cellcolor[HTML]{ffaeae} 41.88\% &  45.03\% & \cellcolor[HTML]{ffd6d6} 43.46\% & \cellcolor[HTML]{ffc9c9} 42.93\% & \cellcolor[HTML]{ffe4e4} 43.98\% & \cellcolor[HTML]{d7ffd7} 47.12\% & \cellcolor[HTML]{d7ffd7} 47.12\% & \cellcolor[HTML]{e1ffe1} 46.60\% & \cellcolor[HTML]{b0ffb0} \textbf{49.21}\% & \cellcolor[HTML]{c4ffc4} 48.17\% & \cellcolor[HTML]{cdffcd} 47.64\% & \cellcolor[HTML]{e1ffe1} 46.60\% & \cellcolor[HTML]{ffe4e4} 43.98\% & \cellcolor[HTML]{d7ffd7} 47.12\% & \cellcolor[HTML]{cdffcd} 47.64\% &  45.03\% & \cellcolor[HTML]{ffe4e4} 43.98\% & \cellcolor[HTML]{ff9393} 40.84\% & \cellcolor[HTML]{ff5d5d} 38.74\% & \cellcolor[HTML]{ff5050} 38.22\% & \cellcolor[HTML]{ff0000} 35.08\% \\
700 & \cellcolor[HTML]{ffd6d6} 43.46\% & \cellcolor[HTML]{ffbbbb} 42.41\% & \cellcolor[HTML]{ebffeb} 46.07\% &  45.03\% & \cellcolor[HTML]{ffe4e4} 43.98\% & \cellcolor[HTML]{d7ffd7} \textbf{47.12}\% & \cellcolor[HTML]{ffe4e4} 43.98\% & \cellcolor[HTML]{ebffeb} 46.07\% & \cellcolor[HTML]{f5fff5} 45.55\% & \cellcolor[HTML]{ebffeb} 46.07\% & \cellcolor[HTML]{d7ffd7} \textbf{47.12}\% & \cellcolor[HTML]{fff1f1} 44.50\% & \cellcolor[HTML]{ebffeb} 46.07\% & \cellcolor[HTML]{fff1f1} 44.50\% & \cellcolor[HTML]{ffbbbb} 42.41\% & \cellcolor[HTML]{fff1f1} 44.50\% &  45.03\% & \cellcolor[HTML]{ff8686} 40.31\% & \cellcolor[HTML]{ff5050} 38.22\% & \cellcolor[HTML]{ff2828} 36.65\% & \cellcolor[HTML]{ff0d0d} 35.60\% \\
\hline
\end{tabular}
}
\label{tab:vstar_deriv_13b}
\end{table}

\begin{table}[]
\caption{Accuracy on $V^*$ for LLaVA-\modelname-13B using random selection method. Native resolution.}
\vspace{1em}
\centering
\resizebox{\textwidth}{!}{
\begin{tabular}{ |c|c|c|c|c|c|c|c|c| }
\hline
Image Size & $\alpha=0.00$ & $\alpha=0.05$ & $\alpha=0.10$ & $\alpha=0.20$ & $\alpha=0.30$ & $\alpha=0.40$ & $\alpha=0.50$ & $\alpha=0.60$ \\ \hline
224 & \cellcolor[HTML]{7aff7a} 49.21\% & \cellcolor[HTML]{c1ffc1} 45.55\% & \cellcolor[HTML]{47ff47} \textbf{51.83}\% & \cellcolor[HTML]{eaffea} 43.46\% & \cellcolor[HTML]{c1ffc1} 45.55\% & \cellcolor[HTML]{ff9f9f} 39.27\% & \cellcolor[HTML]{ff7f7f} 38.22\% & \cellcolor[HTML]{ff4f4f} 36.65\% \\
252 & \cellcolor[HTML]{70ff70} \textbf{49.74}\% &  42.41\% & \cellcolor[HTML]{adffad} 46.60\% & \cellcolor[HTML]{d6ffd6} 44.50\% & \cellcolor[HTML]{8eff8e} 48.17\% & \cellcolor[HTML]{ff7f7f} 38.22\% & \cellcolor[HTML]{ff3f3f} 36.13\% & \cellcolor[HTML]{ff9f9f} 39.27\% \\
280 & \cellcolor[HTML]{8eff8e} \textbf{48.17}\% & \cellcolor[HTML]{c1ffc1} 45.55\% &  42.41\% & \cellcolor[HTML]{f4fff4} 42.93\% & \cellcolor[HTML]{ffcfcf} 40.84\% & \cellcolor[HTML]{ff8f8f} 38.74\% & \cellcolor[HTML]{ff6f6f} 37.70\% & \cellcolor[HTML]{ff9f9f} 39.27\% \\
308 & \cellcolor[HTML]{7aff7a} \textbf{49.21}\% & \cellcolor[HTML]{d6ffd6} 44.50\% & \cellcolor[HTML]{c1ffc1} 45.55\% & \cellcolor[HTML]{eaffea} 43.46\% &  42.41\% & \cellcolor[HTML]{ffdfdf} 41.36\% & \cellcolor[HTML]{ff5f5f} 37.17\% & \cellcolor[HTML]{ff0000} 34.03\% \\
336 & \cellcolor[HTML]{51ff51} 51.31\% & \cellcolor[HTML]{00ff00} \underline{\textbf{55.50\%}} & \cellcolor[HTML]{8eff8e} 48.17\% & \cellcolor[HTML]{c1ffc1} 45.55\% & \cellcolor[HTML]{ffdfdf} 41.36\% & \cellcolor[HTML]{ffbfbf} 40.31\% & \cellcolor[HTML]{ffafaf} 39.79\% & \cellcolor[HTML]{ffbfbf} 40.31\% \\
364 & \cellcolor[HTML]{1eff1e} \textbf{53.93}\% & \cellcolor[HTML]{47ff47} 51.83\% & \cellcolor[HTML]{5bff5b} 50.79\% & \cellcolor[HTML]{c1ffc1} 45.55\% & \cellcolor[HTML]{ffefef} 41.88\% & \cellcolor[HTML]{ff8f8f} 38.74\% & \cellcolor[HTML]{ffafaf} 39.79\% & \cellcolor[HTML]{ff8f8f} 38.74\% \\
392 & \cellcolor[HTML]{28ff28} \textbf{53.40}\% & \cellcolor[HTML]{28ff28} \textbf{53.40}\% & \cellcolor[HTML]{84ff84} 48.69\% & \cellcolor[HTML]{ccffcc} 45.03\% & \cellcolor[HTML]{d6ffd6} 44.50\% & \cellcolor[HTML]{ff4f4f} 36.65\% & \cellcolor[HTML]{ffafaf} 39.79\% & \cellcolor[HTML]{ffbfbf} 40.31\% \\
420 & \cellcolor[HTML]{66ff66} 50.26\% & \cellcolor[HTML]{84ff84} 48.69\% & \cellcolor[HTML]{33ff33} \textbf{52.88}\% & \cellcolor[HTML]{a3ffa3} 47.12\% & \cellcolor[HTML]{84ff84} 48.69\% &  42.41\% & \cellcolor[HTML]{ffcfcf} 40.84\% & \cellcolor[HTML]{ff0f0f} 34.55\% \\
448 & \cellcolor[HTML]{66ff66} 50.26\% & \cellcolor[HTML]{47ff47} \textbf{51.83}\% & \cellcolor[HTML]{ccffcc} 45.03\% & \cellcolor[HTML]{e0ffe0} 43.98\% & \cellcolor[HTML]{b7ffb7} 46.07\% & \cellcolor[HTML]{ffdfdf} 41.36\% & \cellcolor[HTML]{ffcfcf} 40.84\% & \cellcolor[HTML]{ff6f6f} 37.70\% \\
476 & \cellcolor[HTML]{70ff70} \textbf{49.74}\% & \cellcolor[HTML]{99ff99} 47.64\% & \cellcolor[HTML]{adffad} 46.60\% & \cellcolor[HTML]{8eff8e} 48.17\% & \cellcolor[HTML]{a3ffa3} 47.12\% & \cellcolor[HTML]{a3ffa3} 47.12\% & \cellcolor[HTML]{ff9f9f} 39.27\% & \cellcolor[HTML]{ffcfcf} 40.84\% \\
504 & \cellcolor[HTML]{7aff7a} 49.21\% & \cellcolor[HTML]{8eff8e} 48.17\% & \cellcolor[HTML]{70ff70} \textbf{49.74}\% & \cellcolor[HTML]{c1ffc1} 45.55\% & \cellcolor[HTML]{ccffcc} 45.03\% & \cellcolor[HTML]{ff4f4f} 36.65\% &  42.41\% & \cellcolor[HTML]{ff4f4f} 36.65\% \\
532 & \cellcolor[HTML]{8eff8e} 48.17\% & \cellcolor[HTML]{84ff84} 48.69\% & \cellcolor[HTML]{8eff8e} 48.17\% & \cellcolor[HTML]{66ff66} \textbf{50.26}\% & \cellcolor[HTML]{d6ffd6} 44.50\% & \cellcolor[HTML]{f4fff4} 42.93\% & \cellcolor[HTML]{ffbfbf} 40.31\% & \cellcolor[HTML]{ffbfbf} 40.31\% \\
560 & \cellcolor[HTML]{adffad} \textbf{46.60}\% & \cellcolor[HTML]{c1ffc1} 45.55\% & \cellcolor[HTML]{b7ffb7} 46.07\% & \cellcolor[HTML]{b7ffb7} 46.07\% & \cellcolor[HTML]{b7ffb7} 46.07\% &  42.41\% & \cellcolor[HTML]{eaffea} 43.46\% & \cellcolor[HTML]{d6ffd6} 44.50\% \\
588 & \cellcolor[HTML]{ccffcc} 45.03\% & \cellcolor[HTML]{eaffea} 43.46\% & \cellcolor[HTML]{c1ffc1} 45.55\% & \cellcolor[HTML]{84ff84} \textbf{48.69}\% & \cellcolor[HTML]{c1ffc1} 45.55\% & \cellcolor[HTML]{e0ffe0} 43.98\% & \cellcolor[HTML]{d6ffd6} 44.50\% & \cellcolor[HTML]{ccffcc} 45.03\% \\
616 & \cellcolor[HTML]{ff7f7f} 38.22\% & \cellcolor[HTML]{f4fff4} 42.93\% & \cellcolor[HTML]{c1ffc1} \textbf{45.55}\% & \cellcolor[HTML]{d6ffd6} 44.50\% & \cellcolor[HTML]{e0ffe0} 43.98\% & \cellcolor[HTML]{ffdfdf} 41.36\% & \cellcolor[HTML]{ffcfcf} 40.84\% & \cellcolor[HTML]{ff7f7f} 38.22\% \\
644 & \cellcolor[HTML]{ffcfcf} 40.84\% & \cellcolor[HTML]{ffcfcf} 40.84\% & \cellcolor[HTML]{99ff99} \textbf{47.64}\% & \cellcolor[HTML]{ffefef} 41.88\% & \cellcolor[HTML]{ffbfbf} 40.31\% & \cellcolor[HTML]{ffbfbf} 40.31\% & \cellcolor[HTML]{ffafaf} 39.79\% & \cellcolor[HTML]{ff7f7f} 38.22\% \\
672 & \cellcolor[HTML]{ffafaf} 39.79\% &  42.41\% & \cellcolor[HTML]{f4fff4} \textbf{42.93}\% & \cellcolor[HTML]{ffdfdf} 41.36\% & \cellcolor[HTML]{ff9f9f} 39.27\% & \cellcolor[HTML]{f4fff4} \textbf{42.93}\% &  42.41\% & \cellcolor[HTML]{ff5f5f} 37.17\% \\
700 & \cellcolor[HTML]{ffdfdf} 41.36\% & \cellcolor[HTML]{ffbfbf} 40.31\% &  42.41\% & \cellcolor[HTML]{f4fff4} 42.93\% & \cellcolor[HTML]{e0ffe0} \textbf{43.98}\% & \cellcolor[HTML]{ff9f9f} 39.27\% & \cellcolor[HTML]{ffbfbf} 40.31\% & \cellcolor[HTML]{ff1f1f} 35.08\% \\
\hline
\end{tabular}
}
\label{tab:vstar_random_13b}
\end{table}

\begin{figure}
    \centering
    \includegraphics[width=0.95\linewidth]{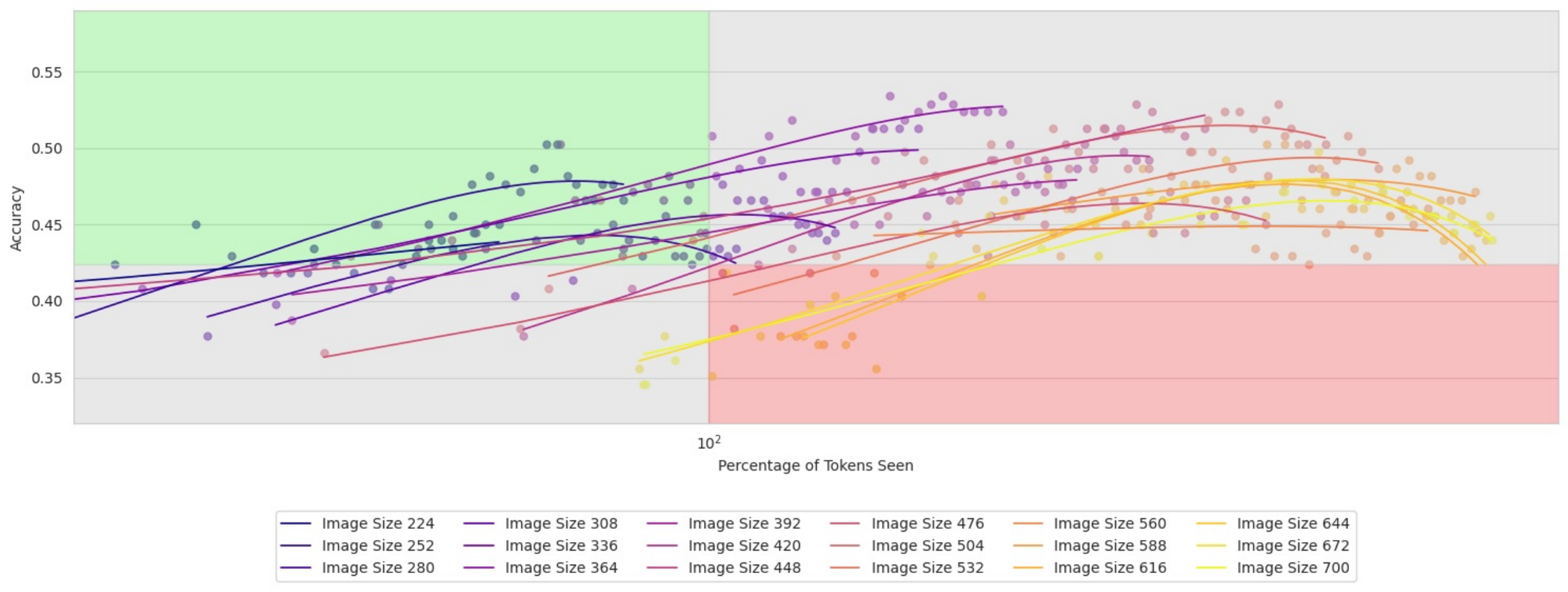}
    \caption{The compute vs. accuracy curves for.     our sweep of $V^*$ with the LLaVA-\modelname-7B model. The $x$-axis is on a logarithmic scale. The green-shaded region highlights experiments where our model \textbf{surpasses the baseline} with \textbf{fewer} visual tokens.}
    \label{fig:enter-label}
\end{figure}

\subsection{Disclosure of Additional Computing Resources}\label{app:evaluation_disclosure}
We did not track the amount of time that our evaluation experiments took, although we plan to update this manuscript with this information once we have re-run the experiments. We had access to two 4x RTX 6000 machines, and one 8x RTX 6000 machine. We variously used compute on these three machines as it became available. Machines are shared between the members of our research group. 
\section{Why We did not to Study QWEN}
The QWEN family of models \citep{yang2024qwen2} includes a vision transformer which is trained from scratch to handle arbitrary resolutions, so the MLP interpolation scheme is not necessary. QWEN implements 2D RoPE \citep{heo_rotary_2024, su2024roformer} which can be interpolated natively by design. We attempted to apply the quadtree selection mechanism to the QWEN vision transformer but we were stopped by particularities in the QWEN model's token merging strategy. In particular they merge adjacent patches before feeding the patches into the vision encoder, which violates the inductive assumptions of the quadtree selection mechanism.
\section{Quadtree Selection Strategy}\label{app:qt_selection}
We found that the directional derivative presented above outperformed more traditional measures like $\max_{x,y}(\,|\partial_x I| + |\partial_yI|\,)$. We do not have an explanation as to why this occurs. It is possible that the averaging strategy is better correlated with patches of interest than looking at the absolute magnitude. We additionally tested variance based methods during the exploratory phase of this project and found that they underperformed our derivative selection strategy.

\subsection{Random Pruning}
Our quadtree implementation works from the root down and decides whether or not to split or not by looking at the split condition. Because we apply quadtree to sub-images of size $2^N\times 2^N$, each sub-image can also be quadtree patchified. To implement random pruning we decide to split a given node with probability $p$, sampled from a uniform distribution.



\section{Detailed Ablations}\label{app:ablations}
\begin{figure}
    \centering
    \includegraphics[width=0.95\linewidth]{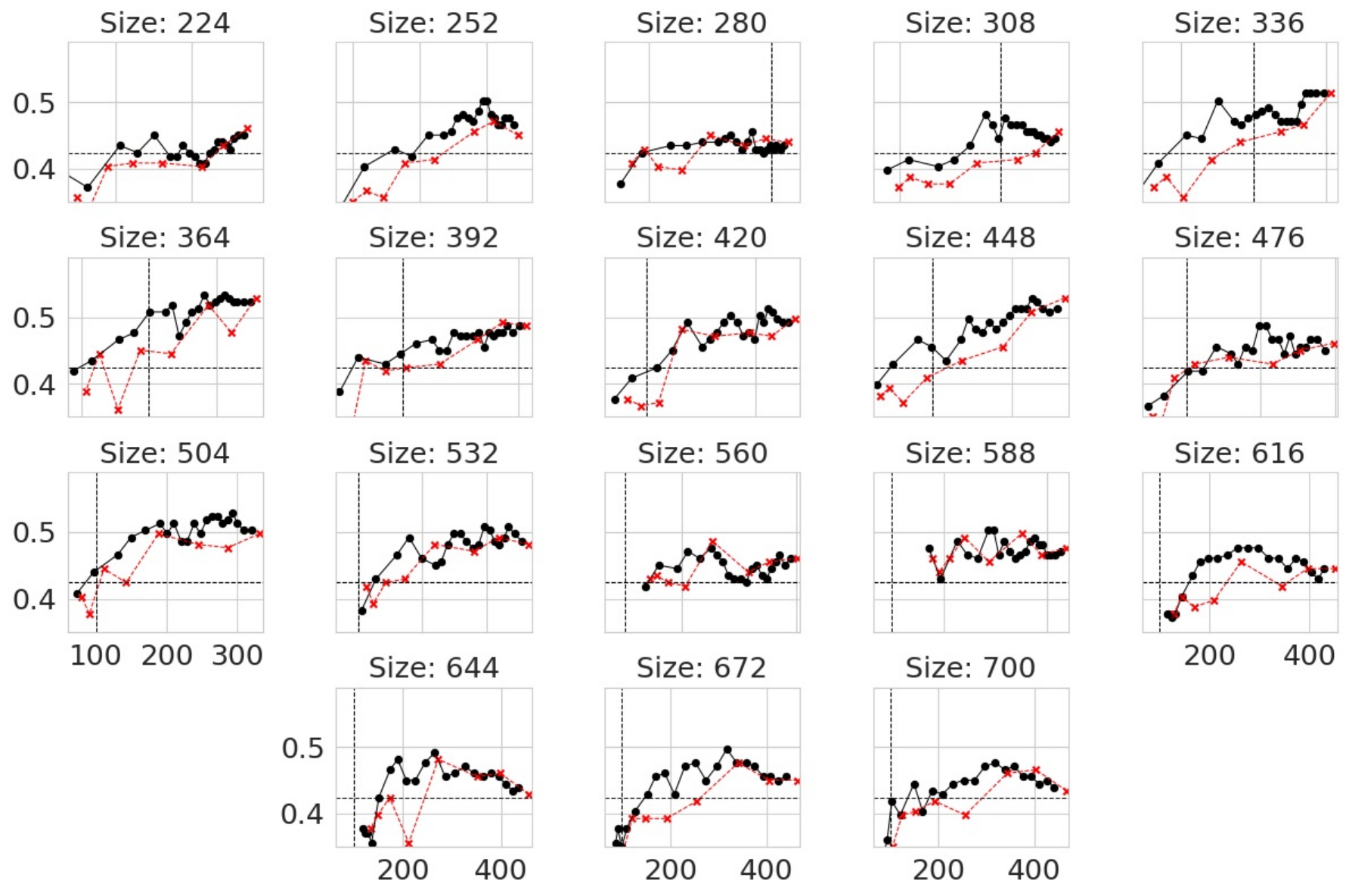}
    \caption{Ablation across a diverse range of image sizes of the \modelname-7B model on the $V^*$ dataset. The Black line is the \modelname performance with the derivative selection strategy and the red line is a random selection strategy. Each random selection trial was only run once.}
    \label{fig:full_ablation_7b_vstar}
\end{figure}

\begin{figure}
    \centering
    \includegraphics[width=0.95\linewidth]{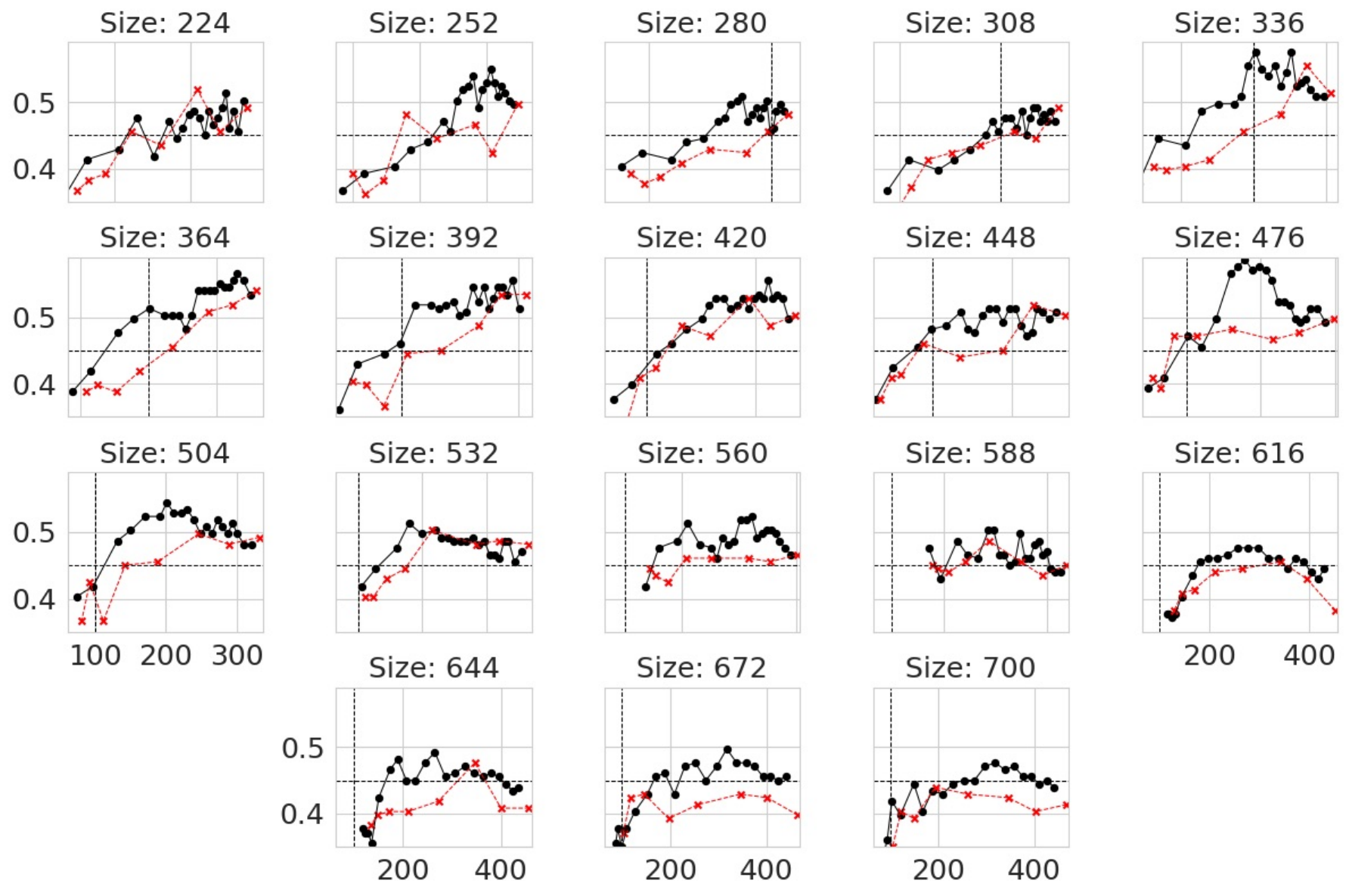}
    \caption{Ablation across a diverse range of image sizes of the \modelname-13B model on the $V^*$ dataset. The Black line is the \modelname performance with the derivative selection strategy and the red line is a random selection strategy. Each random selection trial was only run once.}
    \label{fig:full_ablation_13b_vstar}
\end{figure}

We plot a more comprehensive ablation sweep over $V^*$ than was provided in Figure~\ref{fig:vstar_ablation} above.
 Figure~\ref{fig:full_ablation_7b_vstar} is the ablation for the 7B model on all of the image sizes and values of $\alpha$ that we tested. Figure~\ref{fig:full_ablation_13b_vstar} is the ablation for the 13B model on all of the image sizes and values of $\alpha$ that we tested.

\section{Reproducibility}\label{app:repro}
Our code is available at \url{https://github.com/KyroChi/qlip}. We release all of our code, including training code for the MLP and evaluation code for the trained model. Additionally, we will release our model weights which were used for this paper and also the results that we obtained for this paper.

In our codebase there will be a single command which will
\begin{enumerate}
    \item Run the training script to train an MLP network using the hyperparameters described in this paper.
    \item Run the entire sweep over all of the evaluation benchmarks to reproduce the model results. 
\end{enumerate}
We alternatively include scripts to run reduced evaluations, which are far less computationally expensive than a full parameter sweep. In particular we run the evaluation for the best model that we found for each dataset (Table~\ref{tab:mmbenchmarks})

\section{LLM Statement}
We did not use LLMs in a significant way to aid our research during the completion of this work. Our LLM usage did not extend beyond using code assistants like copilot and for polishing the writing in our manuscript.

\end{document}